\documentclass[10pt,twocolumn,letterpaper]{article}

\usepackage[pagenumbers]{cvpr} 

\usepackage{graphicx}
\usepackage{amsmath}
\usepackage{epstopdf}
\usepackage{amssymb}
\usepackage{booktabs}
\usepackage{makecell}
\usepackage{multirow}
\usepackage{diagbox}
\usepackage[ruled,linesnumbered]{algorithm2e}
\usepackage{enumitem}
\usepackage[capitalize]{cleveref}
\crefname{section}{Sec.}{Secs.}
\Crefname{section}{Section}{Sections}
\Crefname{table}{Table}{Tables}
\crefname{table}{Tab.}{Tabs.}

\def\cvprPaperID{7189} 
\def\confName{CVPR}
\def\confYear{2022}

\begin{document}

\title{Tripartite: Tackle Noisy Labels by a More Precise Partition}

\author{Xuefeng Liang, Longshan Yao, Xingyu Liu, Ying Zhou.\\
School of Artificial Intelligence, Xidian University, China.\\
{\tt\small xliang@xidian.edu.cn}
}
\maketitle

\begin{abstract}
Samples in large-scale datasets may be mislabeled due to various reasons, and Deep Neural Networks can easily over-fit to the noisy labeled data. The key solution is to alleviate the harm of these noisy labels. Many existing methods try to divide training data into clean and noisy subsets in terms of loss values, and then process the noisy labeled data variedly. We observe that a reason hindering a better performance is the hard samples. As hard samples usually have relatively large losses whether their labels are clean or noisy, these methods could not divide them accurately. Instead, we propose a Tripartite solution to partition training data into three subsets: \textit{hard, noisy}, and \textit{clean}. The partition criteria are based on the inconsistent predictions of two networks, and the inconsistency between the prediction of a network and the given label. To minimize the harm of noisy labels but maximize the value of noisy labeled data, we apply a low-weight learning on hard data and a self-supervised learning on noisy labeled data without using the given labels. Extensive experiments demonstrate that Tripartite can filter out noisy labeled data more precisely, and outperforms most state-of-the-art methods on five benchmark datasets, especially on real-world datasets.
\end{abstract}

\section{Introduction}
\label{sec:intro}

Thanks to the large-scale datasets with human precisely annotated labels, DNNs achieve a great success. However, collecting high-quality and extensive data is considerably costly and time-consuming. To alleviate this issue, some cheaper alternatives are often employed, such as web-crawling \cite{mahajan2018exploring}, online queries \cite{blum2003noise,thomee2016yfcc100m,li2017webvision}, crowdsourcing \cite{yan2014learning,yu2018learning} and so on. Unfortunately, they inevitably introduce some noisy labels. Many studies \cite{li2019learning,tanno2019learning,zhang2021understanding,arpit2017closer} have reported that DNNs could easily over-fit to the noises, which significantly degrades their generalization performance.

There have been many efforts to tackle noisy labels. Many of them reach a consensus that the key is to alleviate the impact of noisy labels for network training. Loss-correction based methods \cite{patrini2017making,Goldberger2017TrainingDN,reed2014training,tanaka2018joint} aim to rectify the losses of noisy labeled data in the training stage, but may mistakenly rectify some clean data. Sample-selection based methods \cite{han2018co,li2020dividemix,yu2019does,wei2020combating} tackle noisy labels by partitioning the training data into clean and noisy subsets, then using them for network training in different ways. The mainstream partition criteria are two types: 1) \textit{Small-loss criterion} \cite{han2018co} assumes that the losses of noisy labeled data are significantly higher than those of clean data during training. Therefore, they try to find a threshold, $T$, and select the samples, whose losses $< T$, as clean data. The others are treated as noisy labeled data. 2) \textit{Gaussian Mixture Model} (GMM) \textit{criterion} \cite{li2020dividemix} believes that the statistical distribution of noisy data losses is different from that of clean data losses. It aims to find the probability of a sample being noisy or clean by fitting a mixture model.

\begin{figure*}[htbp]
	\centerline{\includegraphics[scale=0.45]{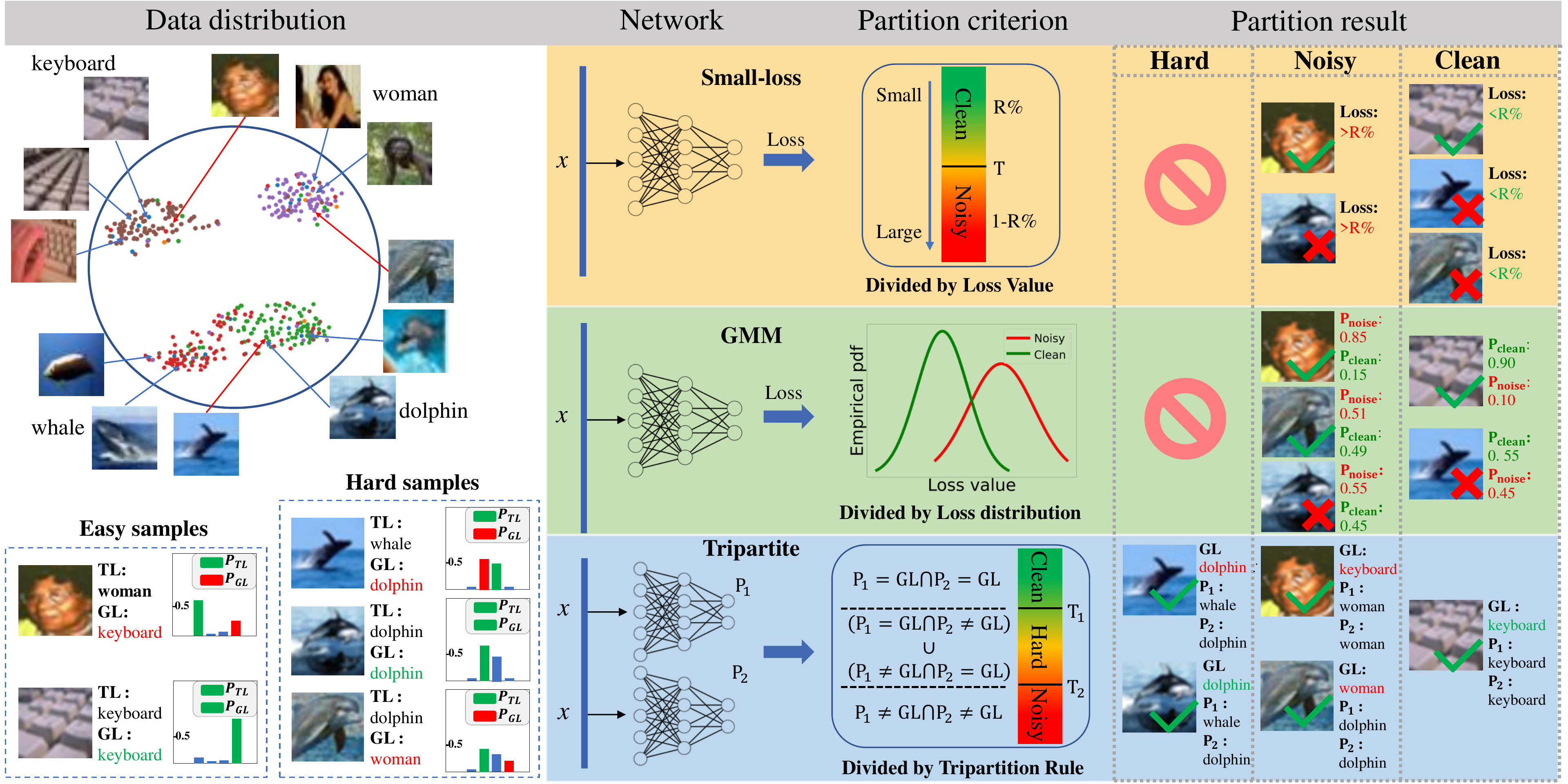}}
	\caption{Comparison with existing partition criteria. From top to bottom are Small-loss criterion, GMM criterion and the proposed Tripartition criterion. TL denotes the true label, GL denotes the given label. $P_1$ and $P_2$ are the predicted labels of inputs. $P_{TL}$ and $P_{GL}$ denote the predicted probabilities of true and given labels, respectively. $P_{clean}$ and $P_{noise}$ denote the predicted probabilities of a sample to be clean and noisy, respectively. In the partition result, the check mark and cross mark denote that the sample is partitioned correctly and incorrectly, respectively. Small-loss criterion sorts losses and selects R\% samples with small losses as clean data for training. GMM criterion finds the probability of a sample being clean or noisy by fitting a mixture model. Neither of them can filter out hard samples. By contrast, Tripartite divides the training set into hard, noisy, and clean subsets according to the relations between $P_1$, $P_2$ and GL.
	}
	\label{fig:partition_narrow}
\end{figure*}

However, we observe that many real-world noisy labels are introduced between similar categories, especially happen among hard samples. For instances, a dolphin is mislabeled as a whale, vice versa, as shown in \cref{fig:partition_narrow}. In this paper, we define “hard samples” as data that distribute close to the decision boundary and are difficult to be distinguished. Our investigation shows, regardless of clean or noisy, the training losses of hard samples are neither small nor significantly different. So, existing sample-selection based methods have two flaws: 1) Low quality of training data partition. They are likely to mistake the hard noisy labeled samples as clean ones solely based on losses, and vice versa. This downgrades the performance of a network because the network will learn certain noisy labels and discard some clean data. 2) Ineffective usage of noisy labeled data. Small-loss criterion often drops the noisy samples which is a waste of valuable information. GMM criterion applies the semi-supervised learning to reuse noisy labeled samples by assigning pseudo-labels to them. But it heavily relies on the discriminative ability of networks. The incorrect relabeling will cause a severe harm to networks as well.

To tackle above problems, we propose a novel method: \textit{Tripartite} that mainly addresses hard samples in real-world datasets. Since hard samples distribute around the decision boundary, predictions of varied networks are often inconsistent at the early training stage. Meanwhile,  the prediction of a network of an easy noisy sample is usually inconsistent with the given label. Based on this observation, Tripartite can divide training data into hard, noisy and clean subsets. It has the advantage of improving the quality of clean and noisy subsets. To effectively use data, we apply a low-weight training strategy for samples in the hard subset, and employ a self-supervised training strategy for samples in the noisy subset without using the given labels. Hence, Tripartite is designed toward minimizing the harm of noisy labels and maximizing the value of noisy labeled data. The extensive experiments on five benchmark datasets demonstrate the superior performance of Tripartite compared with the state-of-the-art (SOTA) methods. Especially, it shows robustness at a wide range of real-world datasets.
The key contributions of our work are threefold.

\begin{itemize}[leftmargin=*]
\item We propose a novel partition criterion, which divides training data into three subsets: hard, noisy, and clean. It alleviates the hard sample selection problem of other criteria, and largely improves the quality of clean and noisy subsets.
\item We design a low-weight training strategy for hard data and a self-supervised training strategy for noisy labeled data, which aim at minimizing the harm of noisy labels and maximizing the value of noisy labeled data.
\item To mimic the noisy label of hard sample in real-world datasets, we create a synthetic class-dependent  label noise on CIFAR datasets, called realistic noise. It flips labels of samples, which are from two different classes, at controlled ratios according to their similarity. The details are shown in Supplementary.
\end{itemize}

\section{Related work}
There have been many studies to address noisy labels. They all try to lower the impact of noisy labels to improve the recognition performance of methods during the training stage. To this end, two different ideas were proposed.

\textbf{Loss correction}. The specific methods include \textit{noise transition matrix} \cite{menon2015learning,natarajan2013learning,patrini2017making,xia2019anchor}, \textit{robust loss functions} \cite{ghosh2017robust,xu2019l_dmi,wang2019imae,zhang2018generalized,wang2019symmetric}, \textit{label correction} \cite{tanaka2018joint,yi2019probabilistic,liu2020early}, etc. \textit{Noise transition matrix} methods construct the label transition matrix to estimate the possibilities of noisy labels transiting among multiple classes. F-correction \cite{patrini2017making} performs forward correction by multiplying the transition matrix with the softmax outputs in the forward propagation. Later, Hendrycks \textit{et al}.\cite{hendrycks2018using} improved the corruption matrix using a small clean dataset. \textit{Robust-loss} methods aim to design loss functions that are robust to noisy labels, such as Mean Absolute Error (MAE) \cite{ghosh2017robust}, Improved MAE \cite{wang2019imae} that is a reweighted MAE, Generalized Cross-Entropy loss (GCE) \cite{zhang2018generalized} that is a generalization of MAE, Symmetric Cross-Entropy \cite{wang2019symmetric} that adds a reverse cross-entropy term to the usual cross-entropy loss. \textit{Label correction} methods try to rectify the noisy labels according to the network predictions during the training. Joint optimization \cite{tanaka2018joint} learns network parameters and infers the true labels simultaneously. PENCIL \cite{yi2019probabilistic} adopts label probability distributions to supervise network learning and update these distributions through back-propagation in each epoch. Inspired by “early learning”, ELR+ \cite{liu2020early} uses the network predictions in the early training stage to correct the noisy labels.

Above methods do not distinguish training data into clean and noisy subsets. They may mistakenly rectify the losses of clean data and introduce new noisy labels into training data. Therefore, sample selection methods were proposed.

\textbf{Sample selection}. These methods try to divide training data into “clean” and “noisy” subsets according to a specific partition criterion. Afterwards, they apply different strategies to train the model on the two subsets separately. The existing partition criteria are mainly based on the training loss, e.g. \textit{small-loss criterion} \cite{han2018co} and \textit{Gaussian Mixture Model} (GMM) \textit{criterion} \cite{li2020dividemix}. \textit{Small-loss criterion} selects training samples with small loss as clean ones by setting a threshold $T$. In particular, MentorNet \cite{jiang2018mentornet} pretrains a teacher network for selecting clean samples to guide the training of a student network. Co-teaching \cite{han2018co} trains two networks where each network selects small-loss samples to feed its peer network for updating parameters. Further more, Co-teaching+ \cite{yu2019does} emphasizes the help of inconsistency to the network, and JoCoR \cite{wei2020combating} emphasizes the consistency of the two networks. However, all these methods only use the data with smaller losses for training without considering the data with larger losses. Meanwhile, finding a feasible $T$ is very challenging. GMM assumes that the distributions of losses of noisy labeled data and clean data follow two normal distributions, respectively. DivideMix \cite{li2020dividemix} employs two networks, each selects data for another in the training, and then applies a semi-supervised method to learn the noisy labeled data. It greatly improves the performance of networks. However, in real-world datasets, the noisy labels often appear between similar classes, such as \textit{whale} and \textit{dolphin}, \textit{oak\_tree} and \textit{maple\_tree} in CIFAR-100. GMM may mistake the hard noisy samples and the hard clean samples because of the little difference of their losses. It leads to a low quality of training data partition. By contrast,  we propose the Tripartition criterion that divides training data into clean, hard, and noisy subsets. A more precise partition can effectively lower the harm of noisy labels for network training.

\section{The proposed method}
To explain the mechanism of Tripartite, we firstly analyze the distributions of hard samples and noisy labeled data, and then derive the logic of data partition in \cref{sec:Distribution}. \cref{sec:Tripartition} details the proposed Tripartition criterion. \cref{sec:strategies} presents the training strategies for data in three subsets. The pseudo-code of Tripartite is shown in \cref{Tripartite}. 

\subsection{Preliminaries}
In our method, we train two networks, denoted by $f(\theta_k), k=\{1,2\}$, which are initialized with different weights. We assume that they could learn different views that include reliable information on noisy training data but produce inconsistent predictions on hard samples. $P_k$ is the predicted label of $f(x,\theta_k)$ on an input $x$. The proposed methods mentioned hereinafter are all based on the predictions of these two networks. Studies \cite{zhang2021understanding,arpit2017closer} report that DNNs tend to learn simple patterns first before memorizing noisy labeled data. And, DNN's optimizations are content-aware, taking advantage of patterns shared by multiple training samples \cite{zhang2021understanding,liu2020early}. Therefore, we assume that networks do not over-fit to the noisy labeled samples in the early learning stage because noisy labels are less related to the data features themselves.

\subsection{Distribution of training data}
\label{sec:Distribution}

\begin{figure}[htbp]
	\centerline{\includegraphics[scale=0.51]{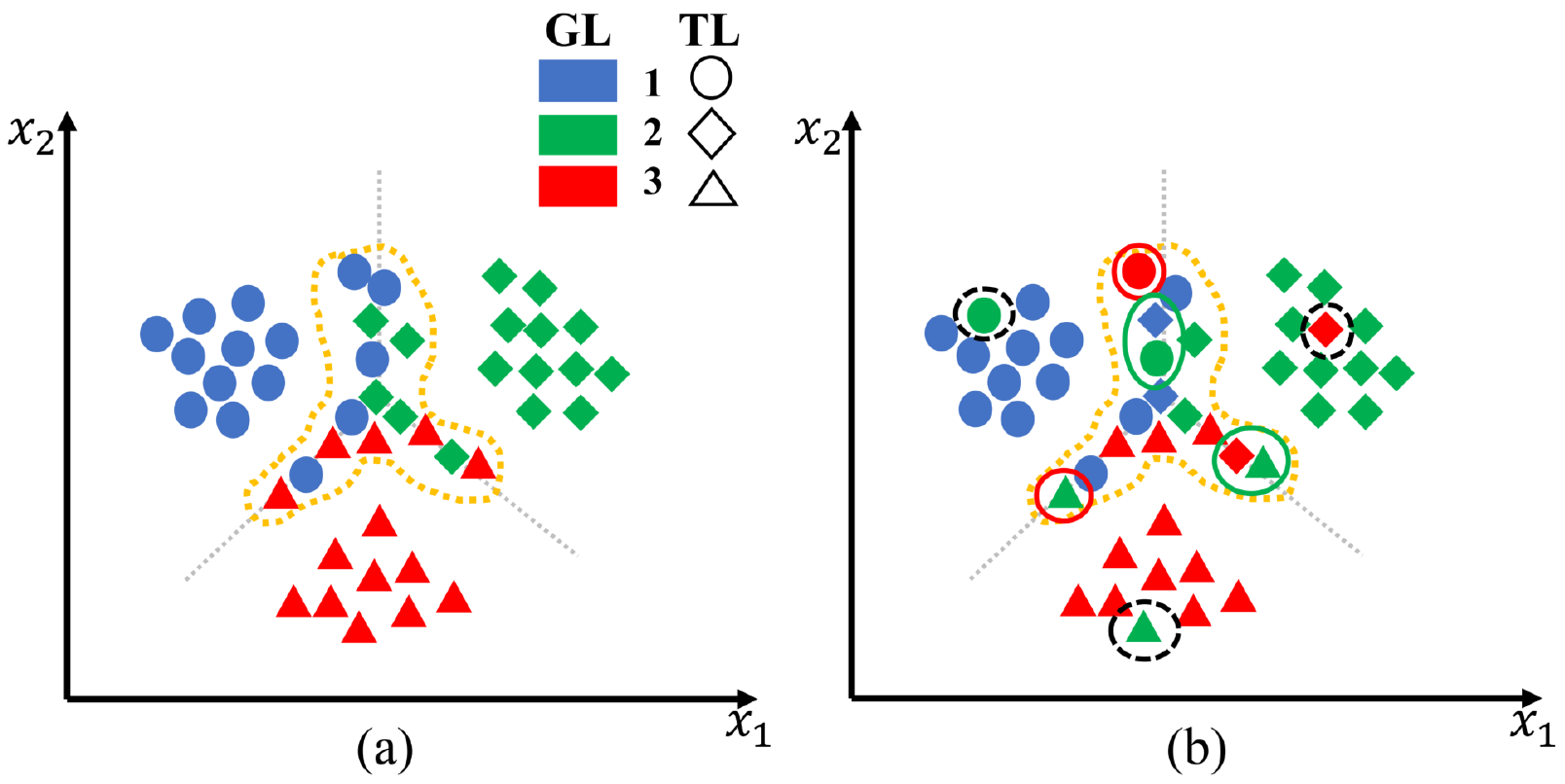}}
    \caption{The distribution of training data. Colors denote the given labels; Shapes denote the true labels. (a) Data with clean labels. Hard samples are grouped by a yellow dotted line. (b) Data with noisy labels. The hard samples with noisy labels from similar/dissimilar classes are circled in green/red. The easy samples with noisy labels are circled in black.}
	\label{fig:example}
\end{figure}

Training data include hard and easy samples. As hard samples share certain features of two or more similar classes, they distribute very close to the decision boundaries. \cref{fig:example} (a) illustrates the data distribution, where hard samples are grouped by a yellow dotted line, the others are easy samples.

For easy samples, there are two cases. 1) \textit{Easy samples with clean labels} whose color and shape are corresponded in \cref{fig:example} (b). As the label information is rather correlated with the sample features, networks can fit these samples in the early training stage. Their losses should be the smallest in the training set. 2) \textit{Easy samples with noisy labels}, circled in black in \cref{fig:example} (b), for instance, a lady is mislabeled as keyboard in \cref{fig:partition_narrow}. As the information provided by the given labels is irrelevant with the features of these samples, they are anomalies in training data. Their losses should be the largest in the training set.

For hard samples, there are three cases. 1) \textit{Hard samples with clean labels}. The label information and sample features are correlated. As distributing close to the decision boundary, they also have some shared features with the similar classes. Therefore, the features of these data are related to both two classes. 2) \textit{Hard samples with noisy labels from similar classes}, circled in green in \cref{fig:example} (b). They distribute close to the decision boundaries and are mislabeled to similar classes. Thus, their features are related to these two classes to some extent. Our observation shows that the losses of both hard samples are greater than the losses of easy clean samples but less than the losses of easy noisy labeled samples. 3) \textit{Hard samples with noisy labels from dissimilar classes}, circled in red in \cref{fig:example} (b), for example, a dolphin is mislabeled as woman. These samples also distribute close to the decision boundaries but are mislabeled to dissimilar classes. The information provided by the given label is irrelevant with the features of these samples, which is analogous to the case of easy noisy samples.



Above analysis illustrates five possible cases of training data, two of easy samples and three of hard samples. In order to improve the quality of data partition and minimize the harm of noisy labels, we then group the five cases into three subsets according to the losses. They are: \textit{hard subset} including hard samples with clean labels, and hard samples with noisy labels from similar classes; \textit{noisy subset} including easy samples with noisy labels, and hard samples with noisy labels from dissimilar classes; \textit{clean subset} including easy samples with clean labels. One can see there exists an order: losses of noisy subset $>$ losses of hard subset $>$ losses of clean subset. In practice, there are two thresholds $T_1$ and $T_2$ in each epoch, where $T_1 < T_2$. Then, training data can be partitioned shown as in \cref{fig:partition_narrow}.

\subsection{Tripartition method and criteria}
\label{sec:Tripartition}
\begin{figure}[t]
	\centerline{\includegraphics[scale=0.35]{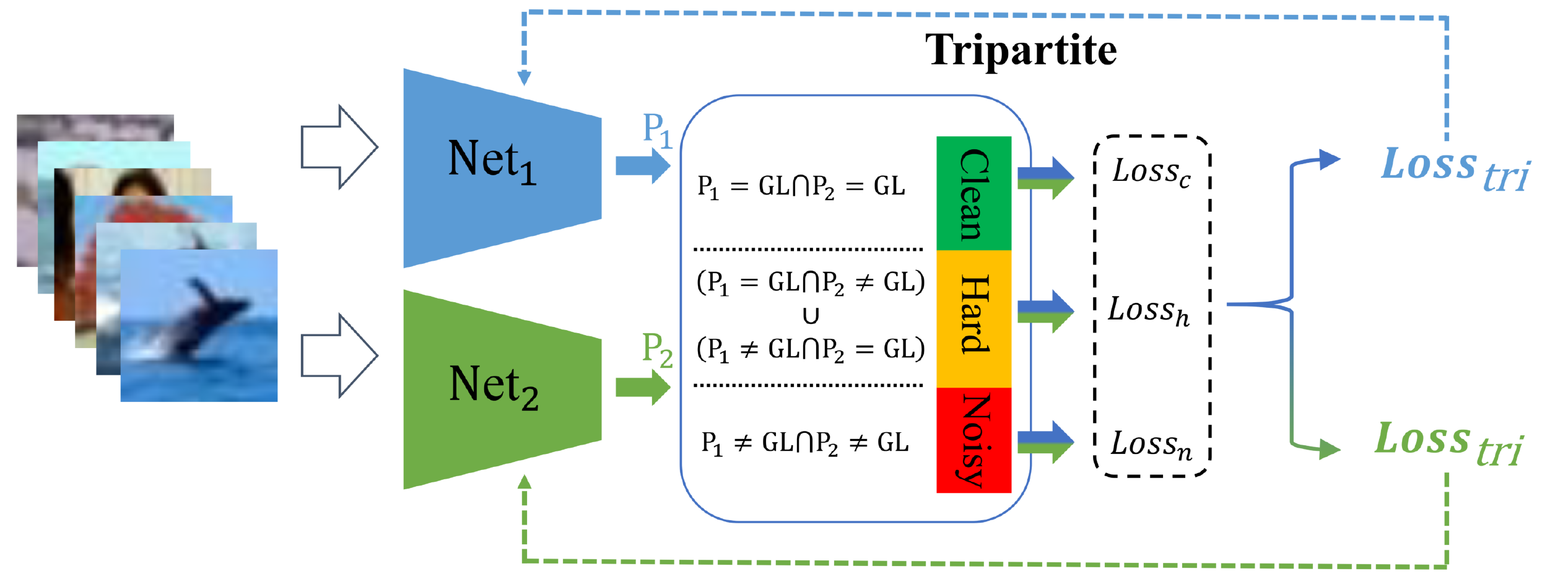}}
	\caption{The logic of our Tripartition method. GL denotes the given label.}
	\label{fig:method}
	\vspace{-0.3cm}
\end{figure}

Theoretically, training data can be partitioned into three subsets according to $T_1$ and $T_2$. Nevertheless, selecting the feasible $T_1$ and $T_2$ is non-trivial. Our investigation shows the predictions of different networks on hard samples usually are inconsistent. There exists a correlation between the losses and the network predictions of training data, which explains why bipartition methods are difficult to handle hard samples. Please refer to Supplementary. Therefore, determining $T_1$ and $T_2$ becomes measuring the consistency of predictions of different networks. Thus, we propose a Tripartition method and the selection criteria. The logic of Tripartite is shown in \cref{fig:method}. We train two networks with different initializations. They give their own predictions in each epoch. Training data are partitioned into three subsets accordingly. \\

\noindent\textbf{Selection criterion for hard subset}
\begin{equation}
	\centering
	\begin{split}
		(P_1 = GL \cap P_2 \neq GL)
		\; \cup \;
		(P_1 \neq GL \cap P_2 = GL)
		\label{eq:partition_eq1}
	\end{split}
\end{equation}


Hard subset mainly includes the hard samples with clean labels and the hard samples with noisy labels from similar classes. The features of these samples and the information provided by labels are somewhat correlated but not very consistent. So, different networks may not give a consistent prediction on such samples, especially at the early training stage. We then apply the inconsistent predictions of two networks, \cref{eq:partition_eq1}, to select them. The losses of selected data should follow $T_1<loss<T_2$. Our experiments indicate that the population of these data will gradually decrease with the increase of the discriminative performance of networks. Please refer to Supplementary for details.\\

\noindent\textbf{Selection criterion for noisy subset}
\begin{equation}
	\centering
	\begin{split}
		P_1 \neq GL \; \cap  \; P_2 \neq GL
		\label{eq:partition_eq2}
	\end{split}
\end{equation}


Noisy subset mainly includes the easy samples with noisy labels and the hard samples with noisy labels from dissimilar classes. As the features of these samples and the information provided by labels are irrelevant and inconsistent, the networks will not fit these data in the early training stage. So, the prediction of a network would not be consistent with the given label. We then apply this inconsistency, \cref{eq:partition_eq2}, to select them. The losses of selected data should follow $loss > T_2$.\\

\noindent\textbf{Selection criterion for clean subset}
\begin{equation}
	\centering
	\begin{split}
		P_1 = GL \; \cap \; P_2 = GL
		\label{eq:partition_eq3}
	\end{split}
\end{equation}

Clean data mainly include the easy samples with clean labels. As the features of these samples and the label information are rather consistent, networks can learn the mapping between data features and labels in the early training stage. So, the predictions of two networks and the given label should be consistent. We then apply this consistency, \cref{eq:partition_eq3}, to select clean data. The losses of selected data should follow $loss < T_1$.

\subsection{Learning strategies for three subsets}
\label{sec:strategies}

We design different learning strategies for three subsets, respectively. Let’s consider a classification problem with
$C$ classes. The training set consists of $n$ examples, $\left \{x^i,y^i\right \}$ is the $i$th input data, $y^i \in \{0,1\}^C$ is a one-hot label of $C$ dimensions vector corresponding to the class of $x^i$. The network maps the input $x^i$ into a $C$-dimensional encoding and feeds it into a softmax function to estimate the probability $p^i$ of $x^i$ to each class.\\

\noindent\textbf{Learning strategy for hard subset}

The hard data subset includes both samples with noisy labels and clean labels. It is difficult to distinguish them. We wish to utilize the valuable information of hard data to improve the discriminative ability of networks, meanwhile, lower the harm of noisy labels in this subset. Therefore, we apply a low-weight cross-entropy loss function for hard data.
\begin{equation}	
	Loss_h=\lambda_h (-\frac{1}{n}\sum_{i=1}^{n}\sum_{c=1}^{C}y_c^i log\left ( p_c^i\right )),
	\label{eq:loss-hard}
\end{equation}

\noindent where, $\lambda_h$ is a weight in the range (0, 1). $p_c^i$ is the estimated probability of $x^i$ to be the class $c$. \\

\begin{figure}[h]
	\centerline{\includegraphics[scale=0.64]{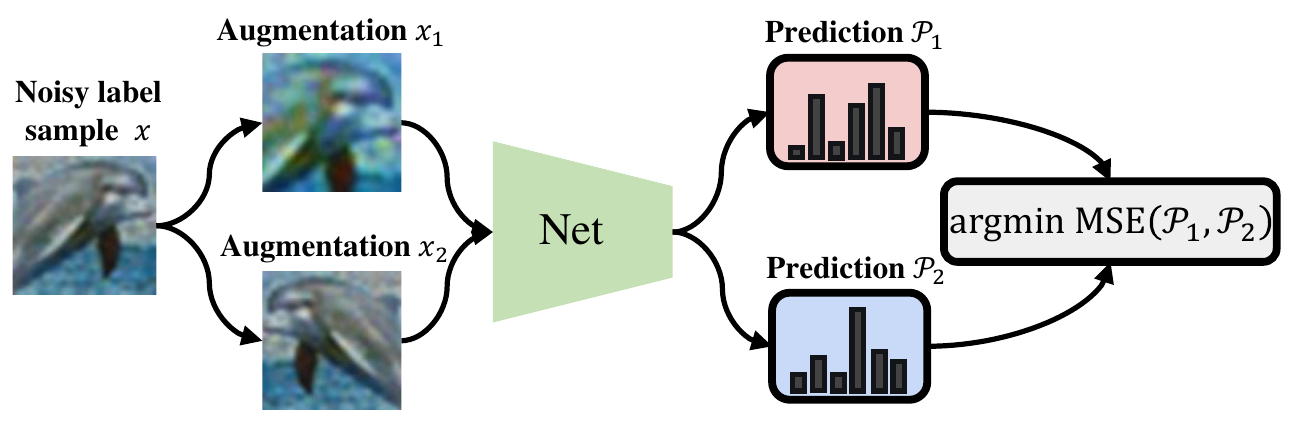}}
	\caption{The logic of our self-supervised strategy. It learns samples in noisy subset without using the given labels.}
	\label{fig:self_learning}
\end{figure}

\noindent\textbf{Learning strategy for noisy subset}

To lower the impact of noisy labels, many works apply semi-supervised learning to replace the given labels by pseudo-labels. However, the quality of pseudo-labels heavily relies on the discriminative ability of networks. New noisy labels may be introduced into training data when the discriminative ability is weak. Instead, we design a self-supervised method to learn these data without using the given labels.

Given a noisy labeled sample $x$ shown in \cref{fig:self_learning}, we randomly augment it by rotation, flip, cropping, desaturation, contrast, blurring, and MixUp \cite{zhang2018mixup}, etc. These augmentations can create many positive sample-pairs, $\langle \bar{x},\hat{x}\rangle$. They are fed into the network, which predicts a probability distribution, $\mathcal{P}=(p_1,p_2,...,p_C)$, for each augmentation. As aiming at learning the common discriminative features from similar samples, we constrain the consistency of two probability distributions $\langle \bar{\mathcal{P}},\hat{\mathcal{P}}\rangle$ using the MSE loss function.

\begin{equation}
	Loss_n=\frac{1}{n}\sum_{i=1}^{n}(\bar{\mathcal{P}}^i-\hat{\mathcal{P}}^i)^2.
	\label{eq:loss-noise}
\end{equation}

Please note that two networks still learn a bit different features because of the random augmentation in each batch. It helps the two networks to give inconsistent predictions on most of the hard samples.

\noindent\textbf{Learning strategy for clean subset}

As Tripartite ensures the high quality of samples in clean subset, we apply the cross-entropy loss for clean data.
\begin{equation}
	Loss_c=-\frac{1}{n}\sum_{i=1}^{n}\sum_{c=1}^{C}y_c^i log\left ( p_c^i\right ).
	\label{eq:loss-clean}
\end{equation}

Finally, the total loss is

\begin{equation}
	Loss_{tri}= Loss_c+ \lambda_n Loss_n+ Loss_h,
	\label{eq:loss}
\end{equation}
\noindent where, $\lambda_n$ is a parameter to control the contribution of noisy subset.

\begin{algorithm}[h]
	\vspace{-0.2cm}
	\caption{Tripartite} \label{Tripartite}
	\begin{small}
		\BlankLine
		\KwIn{Network $f(\theta_k), k=\{1,2\}$, learning rate $\eta$, loss weight $\lambda_h$ of hard subset, loss weight $\lambda_n$ of noisy subset, end epoch of warm up $T_k$, and epoch $T_{max}$, iteration $I_{max}$}
		
		\textbf{Warm up $f(\theta_1),f(\theta_2)$}\\
		\For{$t = T_k,...,T_{max}$}{
			$P_1=f(x,\theta_1)$, $\forall x \in D$\\
			$P_2=f(x,\theta_2)$, $\forall x \in D$\\
			\textbf{Obtain} $Tripart =$ \textbf{Tripartition}$(P_1,P_2,D)$\\
			\For{$k =1,2$}{
				\For{$i=1,2,...,I_{max}$}{
					\textbf{Fetch} mini-batch $D_n$ from $D$\\
					$D_{clean},D_{hard},D_{noise}= Tripart(D_n)$\\
					\textbf{Calculate} $Loss_{c}$ by \cref{eq:loss-clean},$\forall x \in D_{clean}$\\
					\textbf{Calculate} $Loss_{h}$ by \cref{eq:loss-hard},$\forall x \in D_{hard}$\\
					\textbf{Obtain} $x_1,x_2=$ Augment($x$),$\forall x \in D_{noise}$\\
					\textbf{Calculate} $Loss_{n}$ by \cref{eq:loss-noise}\\
					\textbf{Calculate} $Loss_{tri}$ by \cref{eq:loss}\\
					\textbf{Update} $\theta_{k}=\theta_{k}-\eta \nabla Loss_{tri}$
				}
			}
		}
		\KwOut{$\theta_1,\theta_2$}
	\end{small}
	
\end{algorithm}
\vspace{0.2cm}
\section{Experiments}
\subsection{Implementation details and datasets}

\noindent\textbf{Datasets and noise types:}  We verify the effectiveness of our Tripartite extensively on two benchmark datasets (CIFAR-10 \cite{krizhevsky2009learning}, CIFAR-100 \cite{krizhevsky2009learning}) and three real-world datasets (Food-101N \cite{lee2018cleannet}, Clothing1M \cite{xiao2015learning}, WebVision1.0 \cite{li2017webvision}).

Both CIFAR-10 and CIFAR-100 contain 60,000 images of size 32$\times$32. CIFAR-10 includes 10 super-classes. CIFAR-100 has 100 subclasses. We validate two noise settings on them. In symmetric noise setting, we randomly flipping the labels of a given proportion (20\%, 50\%, 80\%) to other labels uniformly in the training set. In realistic noise setting, we designed a new method that ensures the labels are replaced based on the similarities between classes. The details are shown in Supplementary.


Food-101N is a benchmark for visual food classification using the Food-101\cite{bossard2014food} taxonomy. It contains about 310,000 images of food recipes classified into 101 classes. The estimated noise ratio is about 20\%. Clothing1M consists of 1 million training images collected from online shopping websites with labels generated with the surrounding text. Its noise level is about 38.5\%. WebVision1.0 is a large-scale dataset with real-world noisy labels. It contains more than 2.4 million images crawled from the web using the 1,000 concepts in ImageNet ILSVRC2012 \cite{deng2009imagenet}. Its noise level is about 20\%. For ease of comparison with competing methods, we follow the previous work \cite{chen2019understanding} and then do the test on a subset which is the first 50 classes crawled from Google image.

\noindent\textbf{Models and parameters:}  Following previous works, we use an 18-layer PreAct Resnet \cite{he2016identity} for CIFAR-10, CIFAR-100, and train it by the SGD optimizer with a learning rate of 0.02, a momentum of 0.9 and a weight-decay of 0.0005. The batch size is set to 128. We train the network for 300 epochs. The architectures of the two networks used in our method are the same. The learning rate is reduced by a factor of 10 after 150 and 200 epochs. We set the warm-up epochs to 10 for CIFAR-10 and to 30 for CIFAR-100. The most critical parameters in Tripartite are $\lambda_n$ and $\lambda_h$. The $\lambda_n$ is selected from \{1, 10, 20, 40, 60, 80, 100\}, $\lambda_h$ is 0.6 in both CIFAR datasets. 

For real-world datasets, the experiment settings and parameters are the same with CIFAR datasets except training epochs, learning rate and network architectures. We train the network for 100 epochs. The batch size is set to 64. The learning rate is reduced by a factor of 10 after 30 and 60 epochs. We set the warm-up epochs to 5 for Food-101N and Clothing1M, and 30 for WebVsion1.0. Resnet-50 \cite{he2016deep} pre-trained on ImageNet (following previous work \cite{li2019learning}) is for Food-101N and Clothing1M, and inception-resnet v2 \cite{szegedy2017inception} is for WebVsion1.0.

\subsection{Comparison with SOTA methods}
We compare our method with the baseline Standard CE (the backbone + Cross Entropy), and the recent SOTA methods including Bootstrap \cite{reed2015training}, F-correction \cite{patrini2017making}, P-correction (PENCIL) \cite{yi2019probabilistic}, Meta-Learning \cite{li2019learning}, M-correction \cite{arazo2019unsupervised}, Co-teaching \cite{han2018co}, JoCoR \cite{wei2020combating}, DivideMix \cite{li2020dividemix}, ELR+ \cite{liu2020early}, Co-learning \cite{tan2021co}, DSOS \cite{albert2022addressing} and JNPL \cite{kim2021joint}. We repeat following experiments 5 times, and report the average test accuracy of 5 trials for each experiment.\\

\begin{table}[h]
	\vspace{-0.2cm}
	\resizebox{\linewidth}{!}{
	\begin{tabular}{l|ccc||ccc}
		\hline
		\multirow{3}{*}{\diagbox [width=11em,trim=r] {Method}{Dataset \\ \textit{symmetric}}}          & \multicolumn{3}{c||}{\multirow{2}{*}{CIFAR-10}} & \multicolumn{3}{c}{\multirow{2}{*}{CIFAR-100}} \\ 
		& & & & & &\\ \cline{2-7}
		& 20\%     & 50\%    & 80\%    & 20\%     & 50\%    & 80\%    \\ \hline
		Standard CE                 & 86.8     & 79.4    & 62.9    & 62.0       & 46.7    & 19.9    \\
		Bootstrap(2015) \cite{reed2015training}         & 86.8     & 79.8    & 63.3    & 62.1     & 46.6    & 19.9    \\
		F-correction(2017) \cite{patrini2017making}      & 86.8     & 79.8    & 63.3    & 61.5     & 46.6    & 19.9    \\
		Co-teaching+(2019) \cite{yu2019does}       & 89.5     & 85.7    & 67.4    & 65.6     & 51.8    & 27.9    \\
		P-correction(2019) \cite{yi2019probabilistic}      & 92.4     & 89.1    & 77.5    & 69.4     & 57.5    & 31.1    \\
		Meta-Learning(2019) \cite{li2019learning}      & 92.9     & 89.3    & 77.4    & 68.5     & 59.2    & 42.4    \\
		M-correction(2019) \cite{arazo2019unsupervised}       & 94.0       & 92.0      & 86.8    & 73.9     & 66.1    & 48.2    \\
		DivideMix(2020) \cite{li2020dividemix}          & \underline{96.1}     & 94.6    & \underline{93.2}    & 77.3     & \underline{74.6}    & \underline{60.2}    \\
		ELR+(2020) \cite{liu2020early}          & 95.8     & \underline{94.8}    & \textbf{93.3}  & \underline{77.6}   & 73.6  & \textbf{60.8}    \\
		Co-learning(2021) \cite{tan2021co}        & 92.5    & 84.83   & 63.5   & 66.7    & 55.0   & 36.2   \\
		DSOS(2022) \cite{albert2022addressing}         & 92.7    & 87.4   & 62.7   & 75.1    & 66.2   & 38.0   \\ \hline
		Tripartite         & \textbf{96.3}    & \textbf{94.9}   & 92.6   & \textbf{78.7}    & \textbf{74.7}   & 59.8   \\ \hline
	\end{tabular}
	}
	\caption{Accuracy comparisons on CIFAR-10 and CIFAR-100 with symmetric noise. We ran the open code of DSOS, and report the results. The results of ELR+ and Co-learning are from their papers, the rest of results are from \cite{li2020dividemix}. The best results are in bold and the second-best results are underlined.}
	\label{tab:symmetric}
	\vspace{-0.3cm}
\end{table}


\begin{table}[h]
	\centering
	\small
	\resizebox{\linewidth}{!}{
		
		\begin{tabular}{l|ccc||ccc}
		\hline
		\multirow{2}{*}{\diagbox [width=11em,trim=r] {Method}{Dataset\\\textit{realistic}}}            & \multicolumn{3}{c||}{\multirow{2}{*}{CIFAR-10}} & \multicolumn{3}{c}{\multirow{2}{*}{CIFAR-100}} \\ 
		& & & & & &\\ \cline{2-7}
		& 20\%    & 40\%    & 50\%    & 20\%     & 40\%    & 50\%    \\ \hline
		Co-teaching(2018) \cite{han2018co} & 82.7   & 74.4   & 55.8   & 50.3    & 40.9   & 32.5    \\
		JoCoR(2020) \cite{wei2020combating}      & 82.2   & 68.7   & 55.8   & 49.7    & 35.1   & 29.1   \\
		DivideMix(2020) \cite{li2020dividemix}   & 91.1   & 92.3   & \underline{91.8}   &  \underline{76.2}    & 66.1   & 59.5    \\
		ELR+(2020) \cite{liu2020early}   &  \underline{95.1}   &  \underline{92.9}   & 91.2   & 75.8    &  \underline{72.5}   &  \underline{60.6}   \\
		Co-learning(2021) \cite{tan2021co} & 91.6    & 75.5   & 61.7   & 68.5     & 58.5   & 50.2   \\
		DSOS(2022) \cite{albert2022addressing}         & 92.5    & 89.0   & 81.1   & 75.3    & 64.8   & 52.6   \\ \hline
		Tripartite  & \textbf{96.2}   & \textbf{95.9}   & \textbf{94.4}   & \textbf{78.8}    & \textbf{74.9}   & \textbf{68.6}   \\ \hline
	\end{tabular}}

	\caption{Accuracy comparisons on CIFAR-10 and CIFAR-100 with realistic noise. We strictly implemented the competing methods according to their open codes and papers. The backbone is the Pre-ResNet18.}
	\label{tab:real-world}
	\vspace{-0.25cm}
\end{table}

\begin{figure}[t]
	\centerline{\includegraphics[scale=0.253]{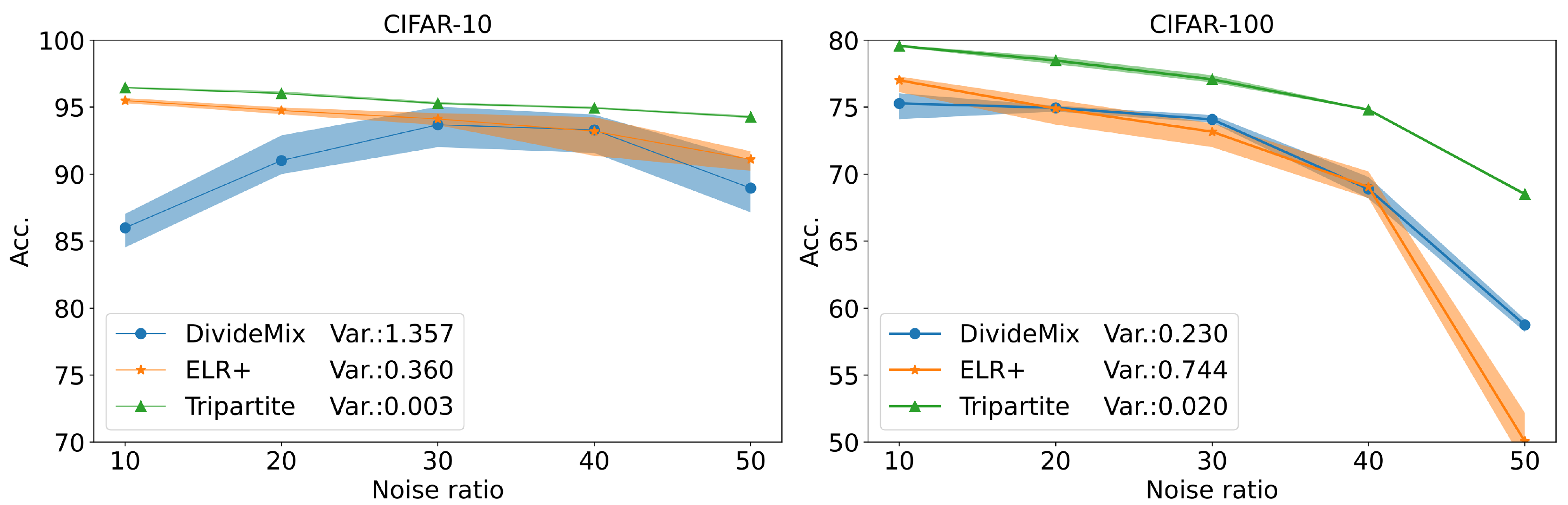}}
    \vspace{-0.2cm}
	\caption{Accuracy comparisons among the top 3 methods on CIFAR with realistic noise. The noise ratio ranges from 10\% to 50\%. The solid lines are the average accuracies. The shaded regions denote the variances.}
	\label{fig:realistic_result}
	\vspace{-0.2cm}
\end{figure}


\noindent\textbf{Results on CIFAR-10 and CIFAR-100}

\cref{tab:symmetric} shows the test accuracy on CIFAR-10 and CIFAR-100 with different levels of symmetric noise ranging from 20\% to 80\%. We can see that Tripartite outperforms competing methods on 20\% and 50\% settings. However, it is slightly worse than the top two methods on 80\% setting. The reason is that most networks have difficulties of learning valuable information during the early training stage when the noise ratio is too high (80\%). This hinders Tripartite to divide training data well. Since Tripartite aims at real-world noisy data, in which the noise ratio over 50\% is unlikely to happen \cite{tan2021co}, we focus on the data with the noise ratio less than 50\%.

\cref{tab:real-world} shows the test accuracy on CIFAR-10 and CIFAR-100 with varied levels of realistic noise. We can see that Tripartite achieves the best across all noise levels. To have a more intuitive comparison, we plot the top 3 methods in \cref{fig:realistic_result}. Compared with DivideMix and ELR+, Tripartite steadily improves the accuracies on every noise ratio, and considerably outperforms them about 3.08\% on CIFAR-10 and 4.54\% on CIFAR-100. Moreover, the variance of Tripartite performances is much lower than those of DivideMix and ELR+. This experiment confirms the superiority of Tripartite on hard noisy labeled data.\\

%
%

\begin{table}[h]
	\begin{minipage}[t]{0.23\textwidth}
		\centering
		\fontsize{8}{9}\selectfont
		\begin{tabular}{l|c}
			\hline
			\multicolumn{2}{c}{Food-101N} \\\hline
			Methods      & Acc.       \\\hline
			Standard CE  & 84.03           \\
			CleanNet \cite{lee2018cleannet} & 83.95\\
			Decoupling \cite{malach2017decoupling}  & 85.53          \\
			Co-teaching \cite{han2018co}  & 61.91          \\
			Co-teaching+ \cite{yu2019does} & 81.61          \\
			JoCoR \cite{wei2020combating}        & 77.94          \\
			Jo-SRC \cite{yao2021jo}  & 86.66          \\
			Co-learning \cite{tan2021co}  & 87.57         \\
			DSOS \cite{albert2022addressing}         & \underline{87.70}   \\ \hline
			Tripartite   & \textbf{88.34} \\ \hline
		\end{tabular}
		\caption{Accuracy comparison on the Food-101N. The results of Jo-SRC  and DSOS are from \cite{yao2021jo}  and \cite{albert2022addressing}. The rest of results are from \cite{tan2021co}.}
		\label{tab:Food-101N}
	\end{minipage}
	\hspace{0.45em}
	\begin{minipage}[t]{0.23\textwidth}
		\centering
		\fontsize{8}{9}\selectfont
		\begin{tabular}{l|c}
			\hline
			\multicolumn{2}{c}{Clothing1M} \\\hline
			Methods       & Acc.         \\\hline
			Standard CE & 69.21         \\
			F-correction \cite{patrini2017making}  & 69.84         \\
			Joint \cite{tanaka2018joint}         & 72.16         \\
			Meta-Learning \cite{li2019learning} & 73.47         \\
			P-correction \cite{yi2019probabilistic}  & 73.49         \\
			DivideMix \cite{li2020dividemix}    & 74.76         \\
			ELR+ \cite{liu2020early}          & \underline{74.81}         \\
			JNPL \cite{kim2021joint}          & 74.15         \\
			DSOS \cite{albert2022addressing}         & 73.63   \\ \hline
			Tripartite    & \textbf{75.23} \\\hline
		\end{tabular}
		\caption{Accuracy comparison on the Clothing1M. The results of JNPL and DSOS are from \cite{kim2021joint} and \cite{albert2022addressing}. The rest of results are from \cite{liu2020early}.}
		\label{tab:Clothing1M}
	\end{minipage}
  \vspace{-0.2cm}
\end{table}

\begin{table}[h]
	\centering
	\fontsize{8}{9}\selectfont
	\begin{tabular}{l|c|c|c|c}
		\hline
		\multirow{2}{*}{\diagbox [width=11em,trim=r] {Method}{Dataset}} & \multicolumn{2}{c|}{WebVision}  & \multicolumn{2}{c}{ILSVRC12}    \\ \cline{2-5}
		&\multicolumn{1}{c|}{Top-1}       & \multicolumn{1}{c|}{Top-5}   &\multicolumn{1}{c|}{Top-1}   &\multicolumn{1}{l}{Top-5}           \\ \hline
		F-correction(2017) \cite{patrini2017making}  & 61.12         & 82.68          & 57.36          & 82.36          \\
		Decoupling(2017) \cite{malach2017decoupling} & 62.54         & 84.74          & 58.26          & 82.26          \\
		D2L(2018) \cite{ma2018dimensionality}         & 62.68         & 84.00            & 57.80           & 81.36          \\
		MentorNet(2018) \cite{jiang2018mentornet}     & 63.00            & 81.40           & 57.80           & 79.92          \\
		Co-teaching(2018) \cite{han2018co}             & 63.58         & 85.20           & 61.48          & 84.70           \\
		Iterative-CV(2019) \cite{chen2019understanding} & 65.24         & 85.34          & 61.60           & 84.98          \\
		DivideMix(2020) \cite{li2020dividemix}           & 77.32         & 91.64    & \underline{75.20}      & \underline{90.84}          \\
		ELR+(2020) \cite{liu2020early}                  & \underline{77.78}   & 91.68          & 70.29          & 89.76          \\
		DSOS(2022) \cite{albert2022addressing}        					   & 77.76         & \underline{92.04}          & 74.36          & 90.80   \\ \hline
		Tripartite                                     & \textbf{78.96} & \textbf{93.20} & \textbf{75.92} & \textbf{92.88} \\\hline
	\end{tabular}
	\caption{Top-1 (Top-5) accuracy comparison on the WebVision1.0 and ILSVRC12. The result of DSOS is from \cite{albert2022addressing}, the rest of results  are from \cite{liu2020early}.}
	\label{tab:webvision}
	\vspace{-0.4cm}
\end{table}

\noindent\textbf{Results on real-world datasets}

\cref{tab:Food-101N,tab:Clothing1M,tab:webvision} list the results on real-world noisy datasets, Food-101N, Clothing1M and WebVision1.0, respectively. We also evaluate the generalization ability of these methods on ImageNet ILSVRC12. Tripartite consistently outperforms competing methods across all large-scale datasets. One can see the performance gains are not very significant on Food-101N and Clothing1M. The possible reason is that images in them often contain more than one subject. Especially, images in Clothing1M usually are full-body shots that include top and bottom clothing but are labeled by either of them. In this case, networks are difficult to learn the discriminative feature. Hence, Tripartite may partition data with some uncertainty.

As WebVision has more and diverse categories, it allows Tripartite to achieve a considerable improvement  than other methods. Please note that ELR+ performs well on WebVision test set, but much worse on ILSVRC12 Top-1 validation set. On the contrary, Tripartite demonstrates its robustness and steady superiority on both WebVision and ILSVRC12.

\subsection{Ablation study}

To verify the effectiveness of our training strategies, the ablation study is conducted on CIFAR-100 in symmetric and realistic settings with varied noise ratios.\\

\noindent\textbf{The quality of training data partition}

\begin{figure}[htbp]
	\centerline{\includegraphics[scale=0.63]{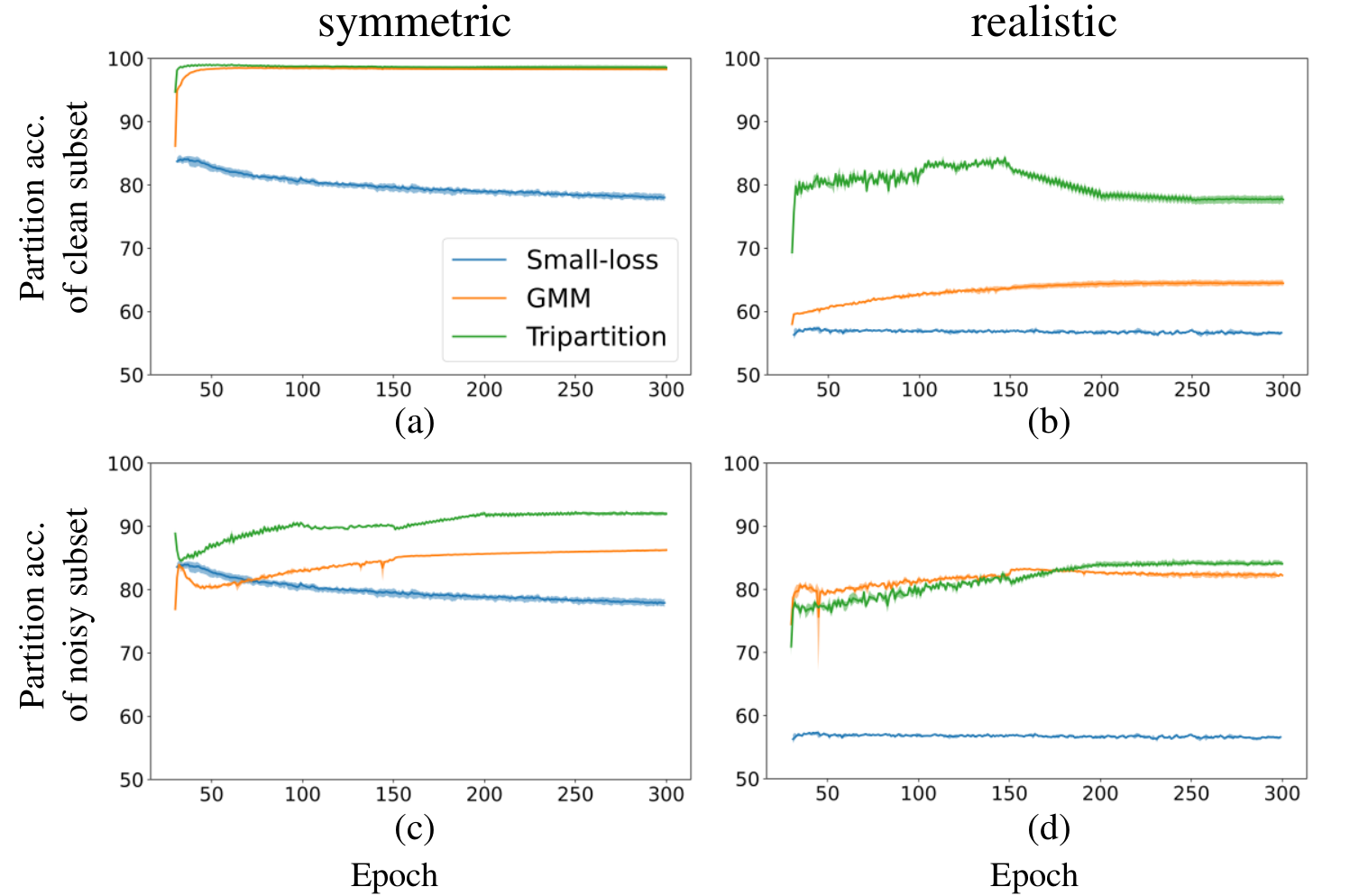}}
	\caption{Comparison of three partition criteria on the CIFAR-100 dataset with symmetric and realistic noise types and 50\% noise ratio. (a) and (b) plot the partition accuracy of clean samples in clean subset against training epochs. (c) and (d) plot the partition accuracy of noisy labeled samples in noisy subset against training epochs.
	}
	\label{fig:partition_result}
\end{figure}

To test the partition accuracies of clean and noisy subsets, we compare Tripartite with \textit{small-loss criterion} and GMM \textit{criterion}. The results are shown in \cref{fig:partition_result}. We can see Tripartite achieves the best partition accuracies on both clean subsets. On the symmetric noise, our Tripartite outperforms the GMM \textit{criterion} slightly. The reason is that the symmetric noises are generated by a symmetric label transfer. The noisy label and clean label can fit the two Gaussian distributions in GMM better in terms of losses. The \textit{small-loss criterion} is the worst because it heavily relies on the selection of R\%. It is worth noting that Tripartite performs better than others, although the population of hard samples with noisy labels from similar classes is smaller on the symmetric type. On the realistic noise, our Tripartite largely outperforms other two criteria on clean subset. As the population of hard noisy labeled data becomes much larger, it narrows the difference between losses of clean and noisy labeled data in the training set, and then further weakens the discriminations of two loss-based criteria.

The comparison of results in \cref{tab:ablation2} and \cref{tab:real-world} also shows that the higher quality of data partition does improve the performance of a model. On realistic noise with a ratio of 50\%, we apply the strategy of co-teaching and JoCoR which trains network without using the noisy subset partitioned by Tripartite. The result (65.04\% in \cref{tab:ablation2}) is much higher than the results (32.5\% and 29.1\% in \cref{tab:real-world}) of Co-teaching and JoCoR. A similar trend can be found on the strategy of DivideMix, which applies the semi-supervised learning on the noisy subset partitioned by Tripartite. The result (65.45\% in \cref{tab:ablation2}) is higher than the results (59.5\% in \cref{tab:real-world}) of DivideMix. \\


\noindent\textbf{Training strategy for hard subset}

\begin{table}[t]
	\fontsize{8}{9}\selectfont
	\caption{Evaluation of $\lambda_h$ in training strategy for hard subset.}
	\centering
	\begin{tabular}{l|c|cccccc}
		\hline
		\multicolumn{2}{c|}{\diagbox [width=8em,trim=r] {Noise type}{$\lambda_h = $}}              & 2     & 1     & 0.8   & 0.6            & 0.4   & 0.2   \\
		\hline
		\multirow{2}{*}{Symmetric} & 20\% & 73.78 & 76.62 & 77.54 & \textbf{78.72} & 78.4  & 76.91 \\
		& 50\% & 60.12 & 66.19 & 70.29 & \textbf{74.73} & 72.92 & 71.06 \\
		\hline
		\multirow{2}{*}{Realistic} & 20\% & 75.65 & 78.07 & 78.35 & \textbf{78.76} & 78.22 & 77.08 \\
		& 50\% & 59.67 & 64.8  & 65.54 & \textbf{68.62} & 67.17 & 65.31  \\
		\hline
	\end{tabular}
	\vspace{-0.1 in}
	
	\label{tab:ablation1}
\end{table}

\begin{table}[h]
	\fontsize{8}{9}\selectfont
	\centering
	\begin{tabular}{l|c|c||c|c}
		\hline
		\multirow{2}{*}{\diagbox [width=8em,trim=r] {Strategy}{Noise type}} & \multicolumn{2}{c||}{Symmetric}   & \multicolumn{2}{c}{Realistic}   \\\cline{2-5}
		& 20\%           & 50\%           & 20\%           & 50\%           \\\hline
		Drop                          & 78.01          & 68.69          & 77.92          & 65.04          \\
		Semi-supervised              & 78.23          & 73.79          & 78.06          & 65.45          \\
		Self-supervised              & \textbf{78.72} & \textbf{74.73} & \textbf{78.76} & \textbf{68.62} \\\hline
	\end{tabular}
	\caption{Evaluation of training strategy for noisy subset.}
	\label{tab:ablation2}
	\vspace{-0.4cm}
\end{table}
The weight $\lambda_h$ is a key parameter for training hard subset. Our motivation is to lower the harm of noisy labels, therefore, we test the $\lambda_h$ in the range of (0,2], and list results in \cref{tab:ablation1}. Obviously, $\lambda_h=2$ reaches the worst result because it amplifies the harm. When $\lambda_h=0.6$, we have the best performance. However, $\lambda_h=0.2$ results in a worse result. The possible reason is that there exist some hard clean  samples, which are critical to the performance of the network. If $\lambda_h$ is too low, the network learns little valuable information from them.\\

\noindent\textbf{Training strategy for noisy subset}

To evaluate the effectiveness of training strategy for noisy subset, we test three strategies and report the results in \cref{tab:ablation2}. One can see that our self-supervised strategy achieves the best because it is label free and learns the common features from similar samples. It can minimize the harm of noisy labels and fully use these data. The $\lambda_n$ is not highly sensitive to Tripartite, see details in  Supplementary. The semi-supervised strategy is worse because it generates pseudo-labels for noisy labeled data to reuse them. The quality of a pseudo-label heavily relies on the performance of the networks. Some incorrect pseudo-labels will degrade the final performance. The strategy of dropping noisy labeled data performs the worst, because it does not use these data.

\section{Conclusion}
Existing loss-based methods divide training data into clean and noisy subsets, but may mistake the hard noisy labeled samples and the hard clean samples. To further improve the quality of two subsets, we propose a novel method, Tripartite, to mainly addresses hard samples. Unlike bipartition, Tripartite partitions training data into three subsets: \textit{hard}, \textit{noisy}, and \textit{clean}. The partition criteria are based on the inconsistent predictions of two networks, and the inconsistency between the prediction of a network and the given label. To minimize the harm of noisy labels but maximize the value of noisy labeled data, Tripartite applies a low-weight strategy on hard data and a self-supervised strategy on noisy labeled data. The extensive experiments demonstrate that Tripartite outperforms SOTA methods and provides a new perspective of data partition and data utilization.


{\small
\bibliographystyle{ieee_fullname}
\bibliography{egbib}
}

\end{document}


	\title{Supplementary Materials 
	
	}
	\maketitle

	\section{The design of realistic noisy}
	\label{sec:intro}
	
	\begin{table} [b]
		\vspace{-.2in}
		\small
		\centering
		
		\setlength{\tabcolsep}{6mm}{
			\centerline{
				\begin{tabular} {c c}
					\hline
					Dataset	& Acc.(\%) \\
					\hline
					CIFAR-10&	94.67\\
					CIFAR-100&	78.85\\
					\hline
		\end{tabular}}}
		\caption{Accuracy of ResNet50 on CIFAR-10/CIFAR-100.}
		\label{tab:result}
		\vspace{-0.2in}
	\end{table}
	
	In the reality, the noisy labels are often introduced on ambiguous data. Thus, the artificial noise types are unlikely to happen in the real world. For instance, the symmetric noise randomly replaces the labels of a certain percentage of the training data by other labels. Although many existing methods can achieve a remarkable performance on such artificial noise types, they may not perform well on the real-world noisy datasets. Instead, we propose a new noise type on CIFAR datasets to mimic the real-world noisy datasets called realistic noise. It flips labels in class-pairs at a varying ratio according to the similarity between two classes. The detailed procedure to generate the noise transition matrix of realistic noise is shown below.
	
	\textbf{Step 1.} We firstly train a ResNet50 on CIFAR-10/CIFAR-100 until it converges. \cref{tab:result} shows the test accuracy of the trained ResNet50 on both CIFAR-100 and CIFAR-10 datasets. We then consider the weight vector between the last fully connected layer and a node (a class) in the output layer as the prototype of the class \cite{Bukchin2021FinegrainedAC}.
	
	\textbf{Step 2.} We pair all prototypes into $N$ class-pairs $(c_i,c_j )$, $i \neq j$, then compute their cosine similarities and sort them in descending order. We select the top $K$ pairs, and flip some labels in each pair, where $K$ is set to 10 for CIFAR-10 (\cref{tab:des}) and 60 for CIFAR-100 (\cref{tab:class}).
	
	\textbf{Step 3.} The $K$ pairs is partitioned into three similarity levels. Each level has a weight $w$, which is proportional to the similarity of the class-pair. Higher $w$ is, more labels will be flipped. $w$ is selected from $\{0.9, 0.6, 0.3\}$ for CIFAR-100, and $\{0.9, 0.8, 0.7\}$ for CIFAR-10. As more similar pairs are selected from CIFAR-100, the weight step between different levels become larger.
	
	\textbf{Step 4.} To construct the noise transition matrix, we temporally fill in the positions of the top $K$ class-pairs with the corresponding weights ${w^{i,j}}$, where $i$ and $j$ denote the $ith$ row and $jth$ column in the matrix. Other positions are filled in with zero. In order to let the sum of all weights in each row to be 1, the $w^{i,j}$ is replaced by
	
	\begin{equation}
		\hat{w}^{i,j} = \frac {w^{i,j}}{\sum_{j=1}^{n} w^{i,j} },
	\end{equation}
	where $n$ is the length of columns.
	
	\textbf{Step 5.} Given a noise ratio $r$, we fill in the diagonal of the matrix with $(1-r)$ and multiply the other elements by $r$.
	
\begin{figure}[h]
		\centering
		\begin{subfigure}{0.48\textwidth}
			\centering\includegraphics[width=6.7cm]{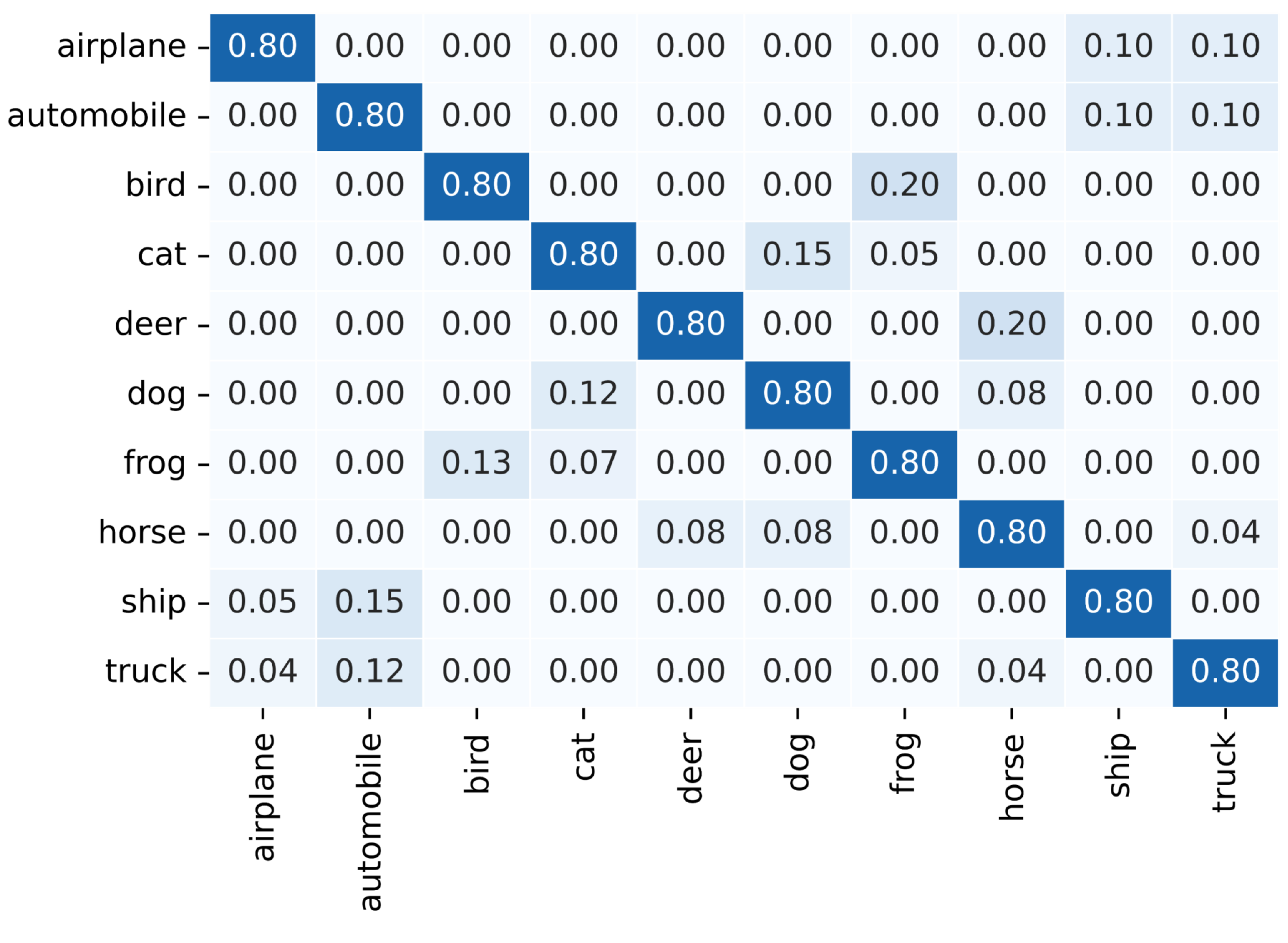}
		\end{subfigure}
		\begin{subfigure}{0.48\textwidth}
			\centering\includegraphics[width=6.7cm]{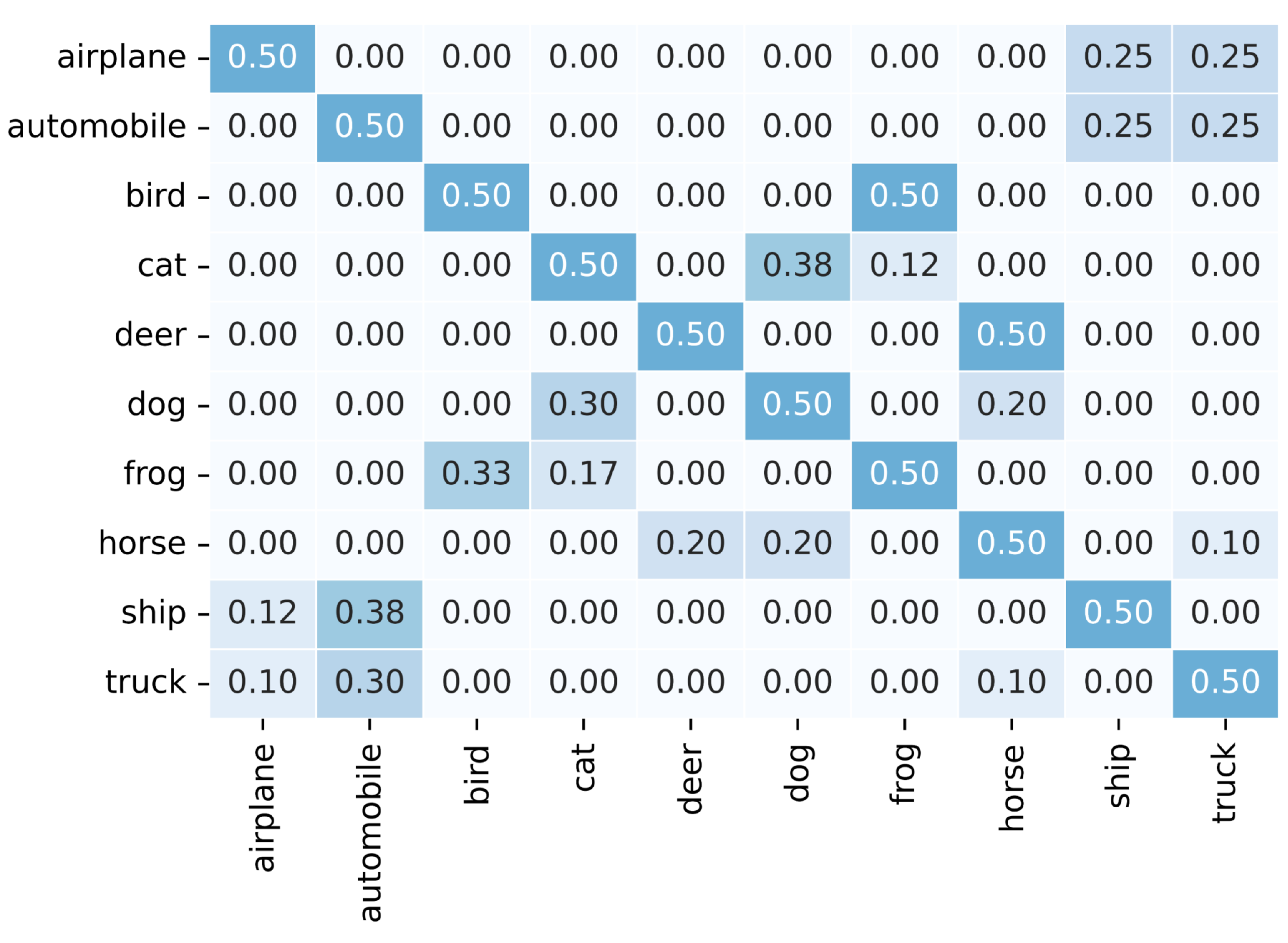}
			
		\end{subfigure}
		\caption{The generated noise transition matrices of CIFAR-10 with noise ratios 20\% and 50\%.}
		\label{fig:matirx}
	\end{figure}

	\cref{fig:matirx} shows the generated noise transition matrices of CIFAR-10 with noise ratios 20\% and 50\%.

\begin{figure}[h]	
		\begin{center}
			\includegraphics[width=9.3cm]{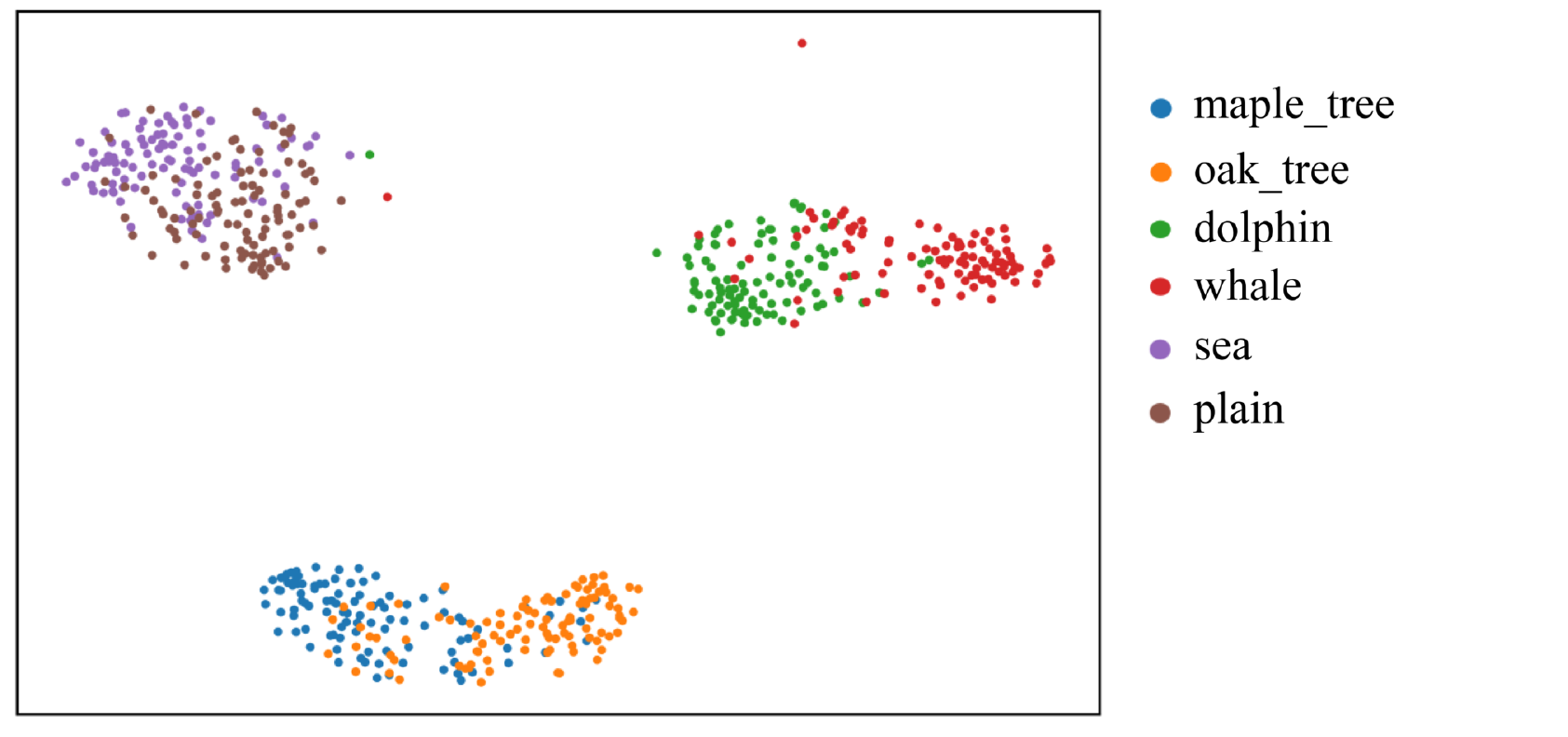}
		\end{center}
		\caption{The visualization of the feature distributions of three similar class-pairs in CIFAR-100 using t-SNE \cite{Maaten2008VisualizingDU}.}
		\label{fig:tsnevisualization}
	\end{figure}

	\cref{fig:tsnevisualization} visualizes the feature distributions of three similar class-pairs in CIFAR-100 using t-SNE. We train a ResNet50 for 200 epochs on CIFAR-100 without noisy labels, and use the output of the last convolutional layer as the feature of each sample. We can see that many hard data distribute close to the decision boundary between two similar classes.
	
	\begin{figure}[h]	
		\begin{center}
			\includegraphics[width=8cm]{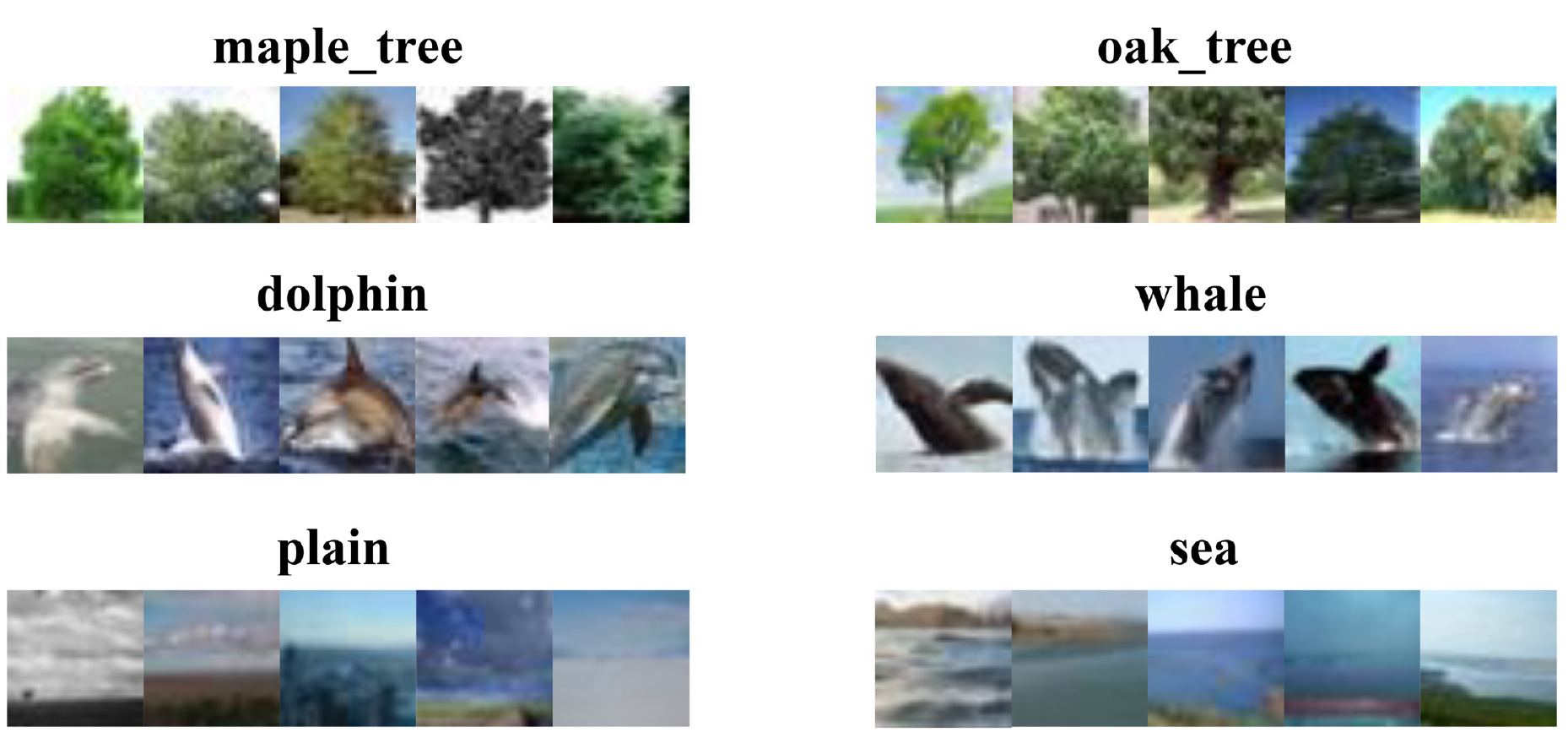}
		\end{center}
		\caption{Some samples selected from the top 3 similar class-pairs in CIFAR-100.}
		\label{fig:cifar100}
	\end{figure}


	\begin{table}[h]
		\centering
		\small	
		\begin{tabular}{ll|c}
			\hline
			\multicolumn{2}{c|}{Similar classes}	&Similarity\\ \hline
			automobile & truck & 0.075 \\
			automobile & ship  & 0.062 \\
			cat        & dog   & 0.062 \\
			dog        & horse & 0.054 \\
			deer       & horse & 0.052 \\
			bird       & frog  & 0.046 \\
			airplane   & ship  & 0.044 \\
			horse      & truck & 0.025 \\
			cat        & frog  & 0.021 \\
			airplane   & truck & 0.016 \\
			\hline
		\end{tabular}
		\caption{The top 10 similar class-pairs in descending order from CIFAR-10. The number is the cosine similarity.}
		\label{tab:des}
	\end{table}

	\section{The motivation of Tripartition criterion}
	\cref{fig:loss_sym,fig:loss_real,fig:loss_web,fig:loss_food,fig:loss_clothing} show the distribution of normalized losses of all samples from CIFAR-100 and WebVision, Food-101N, Clothing1M datasets partitioned by our Tripartition criterion. We can see that the losses of samples in clean, hard and noisy subsets gradually increase. This observation is consistent with our hypothesis in the paper, i.e. losses of clean subset $<$ losses of hard subset $<$ losses of noisy subset. To filter out the hard subset, we need to determine a feasible range [$T_1$, $T_2$].
	
	One can see the loss distributions of three subsets gradually change with the progressing of training. It is nontrivial to determine a feasible range [$T_1$, $T_2$] dynamically. Instead, our proposed Tripartition criterion wisely addresses this problem. It can dynamically partition the data and result in more accurate clean and noisy subsets.
	
	In addition, \cref{fig:loss_real,fig:loss_web} show that the loss distribution of realistic noise is more similar to the one of WebVision dataset compared to symmetric noise. Especially, the mean and variance of loss distributions of noisy subset and hard subset partitioned by Tripartition criterion are close to the ones of WebVision. This demonstrates that the proposed realistic noise is more consistent with the real-world noise type from the perspective of the loss distribution.

	\section{The changes of hard subset}
	
\begin{figure}[h]	
		\begin{center}
			\includegraphics[width=8.5cm]{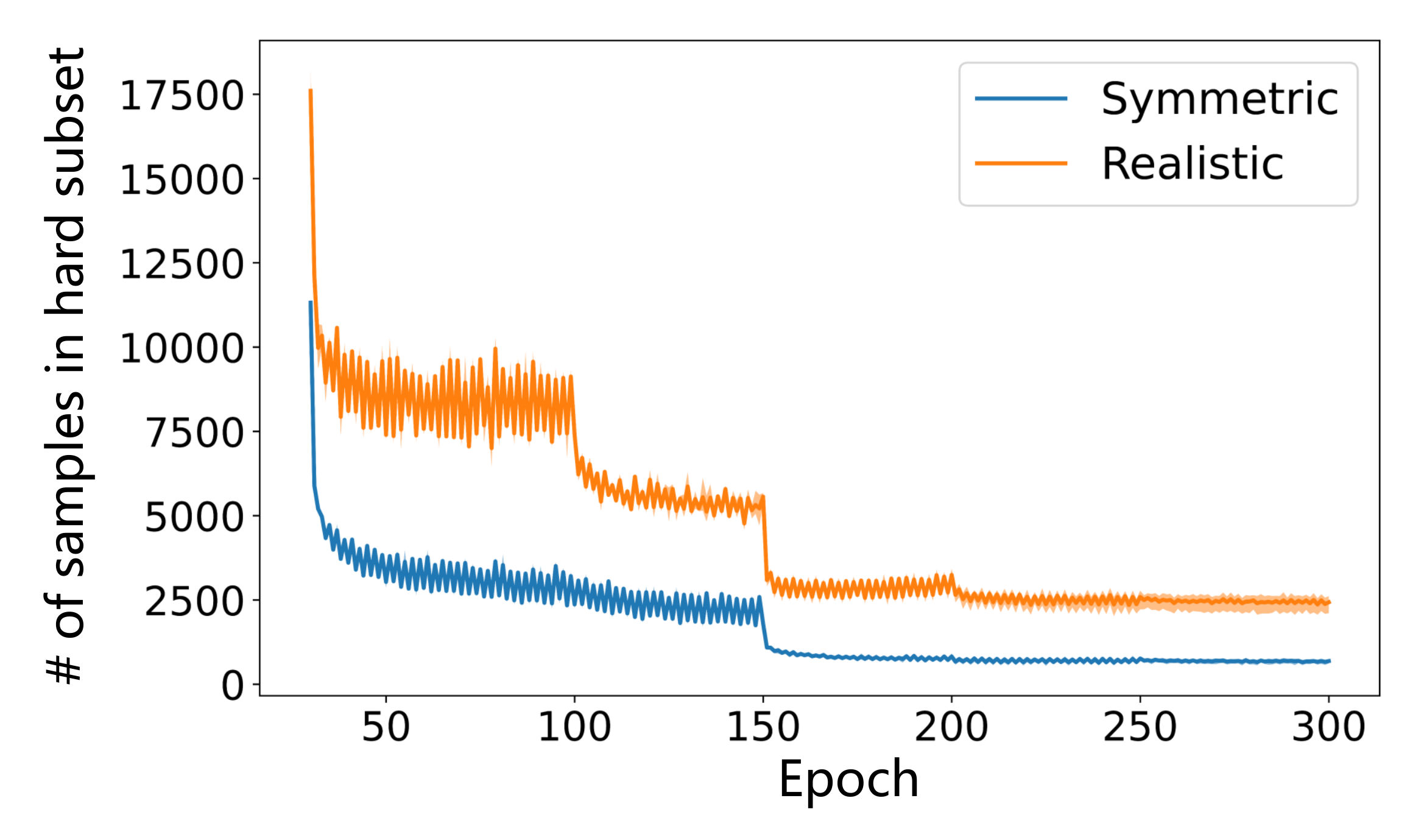}
		\end{center}
		\caption{The changes of population of hard subset in training stage. }
		\label{fig:loss1}
		\vspace{-.05in}
	\end{figure}

\begin{figure*}[t!]	
		\begin{center}
			\includegraphics[width=13cm]{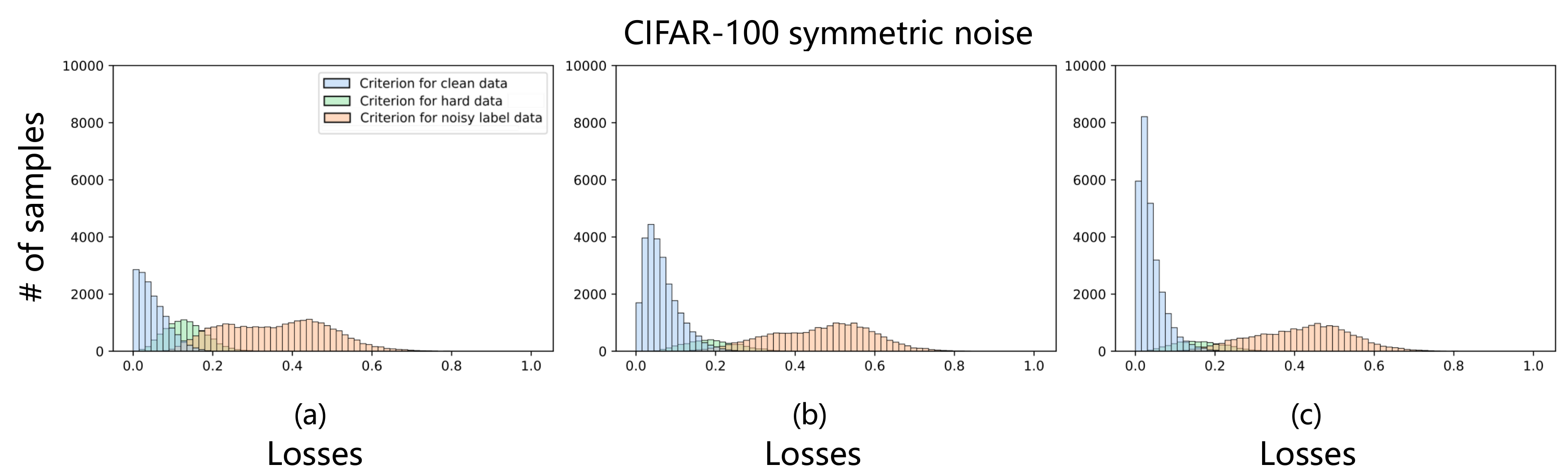}
		\end{center}
		\vspace{-.25in}
		\caption{The distribution of normalized losses of all samples from CIFAR-100 partitioned by our Tripartition criterion. A PreActResNet-18 is trained with 30\% symmetric noise. (a) The distribution at the end of the tenth epoch. (b) The distribution at the end of the fifieth epoch. (c) The distribution of epochs at the optimal accuracy. }
		\label{fig:loss_sym}
		\vspace{-.05in}
	\end{figure*}

\begin{figure*}[t!]	
		\vspace{-.11in}
		\begin{center}
			\includegraphics[width=13cm]{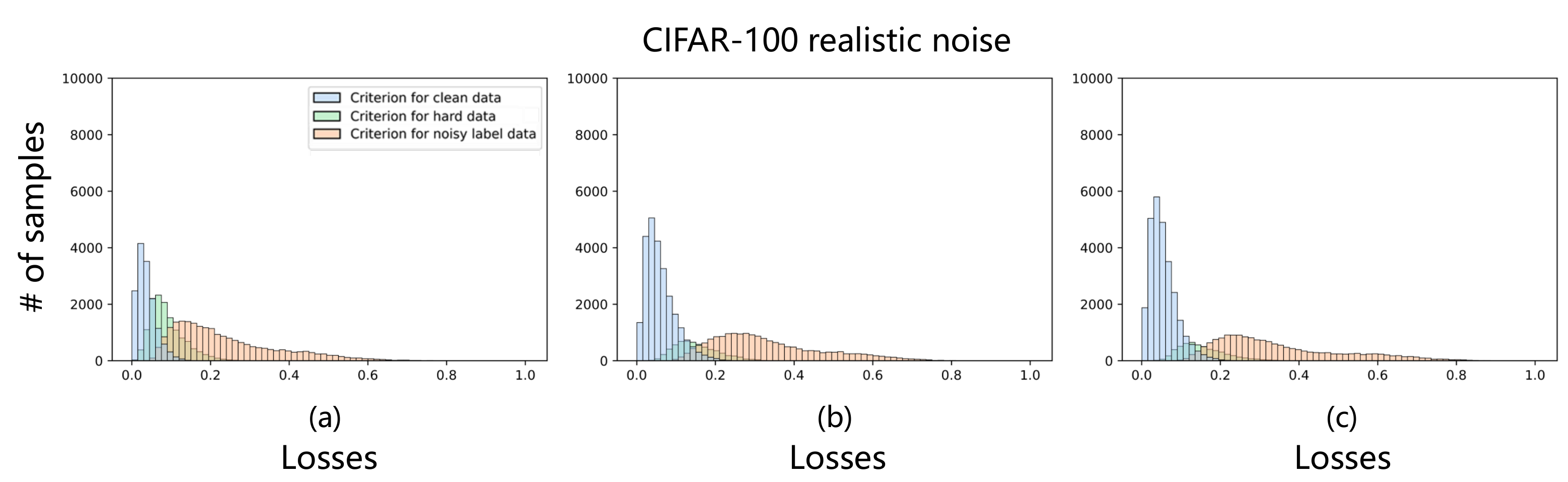}
		\end{center}
		\vspace{-.25in}
		\caption{The distribution of normalized losses of all samples from CIFAR-100 partitioned by our Tripartition criterion. A PreActResNet-18 is trained with 30\% realistic noise. (a) The distribution at the end of the tenth epoch. (b) The distribution at the end of the fifieth epoch. (c) The distribution of epochs at the optimal accuracy.}
		\label{fig:loss_real}
		\vspace{-.05in}
	\end{figure*}

\begin{figure*}[t!]	
		\vspace{-.11in}
		\begin{center}
			\includegraphics[width=13cm]{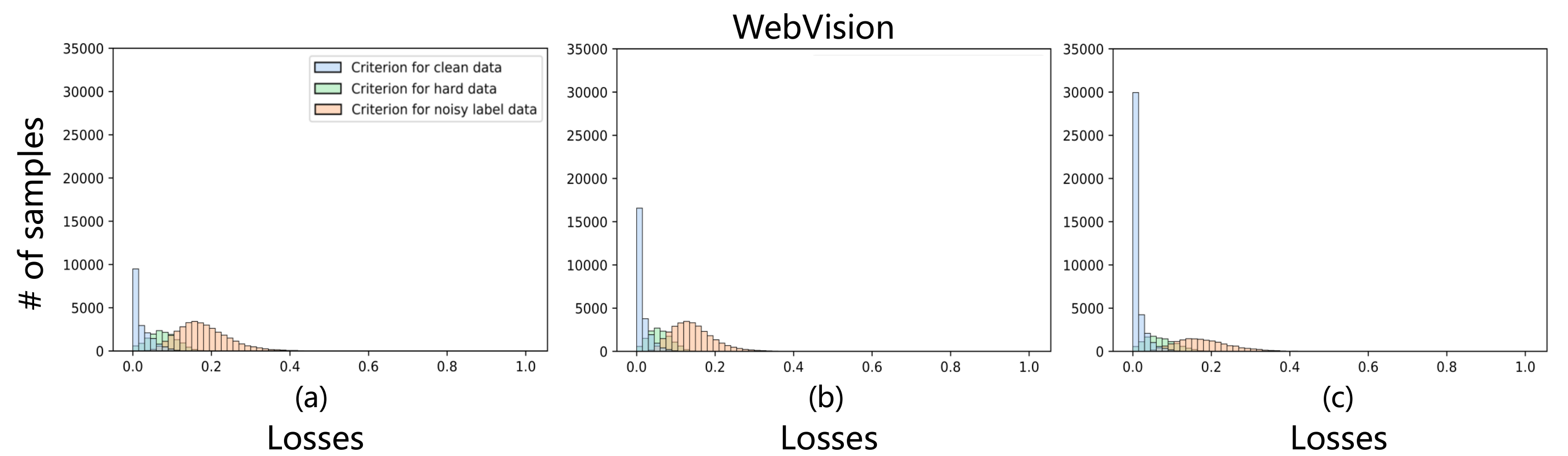}
		\end{center}
		\vspace{-.25in}
		\caption{The distribution of normalized losses of all samples from WebVision partitioned by our Tripartition criterion. (a) The distribution at the end of the sceond epoch. (b) The distribution at the end of the fourth epoch. (c) The distribution of epochs at the optimal accuracy.}
		\label{fig:loss_web}
		\vspace{-.4in}
	\end{figure*}

\begin{figure*}[t!]	
	\begin{center}
		\includegraphics[width=13cm]{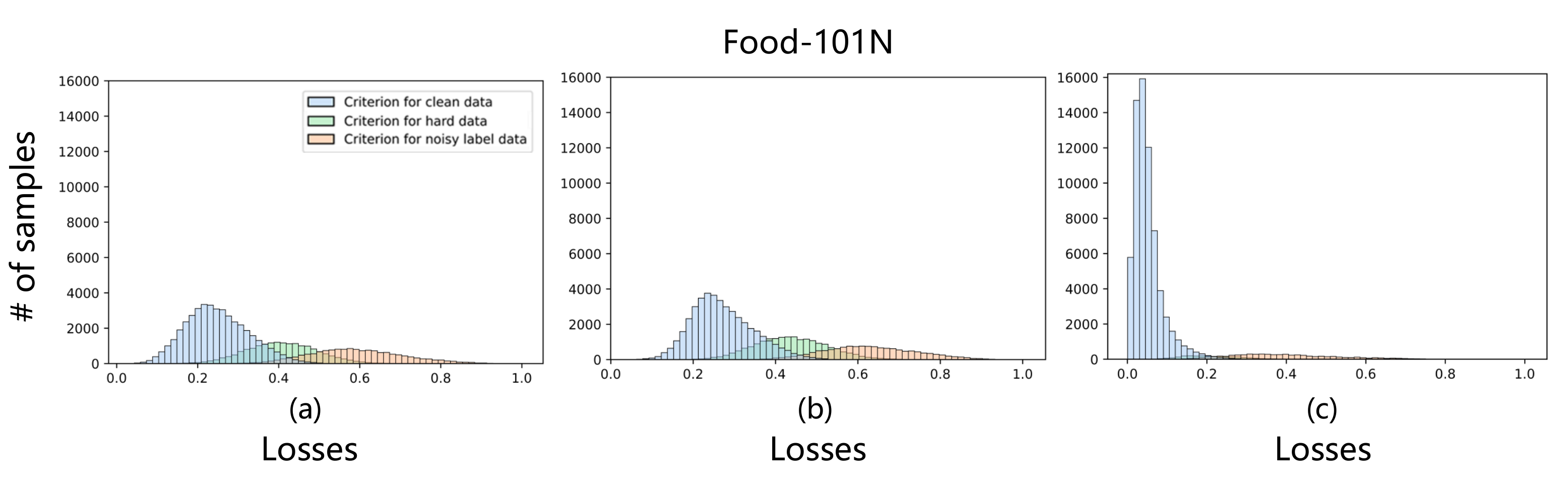}
	\end{center}
	\caption{The distribution of normalized losses of all samples from Food-101N partitioned by our Tripartition criterion. The model is pre-trained on ImageNet. (a) The distribution at the end of the second epoch. (b) The distribution at the end of the fifth epoch. (c) The distribution of epochs at the optimal accuracy.}
	\label{fig:loss_food}
	\end{figure*}

\begin{figure*}[t!]	

	\begin{center}
		\includegraphics[width=13cm]{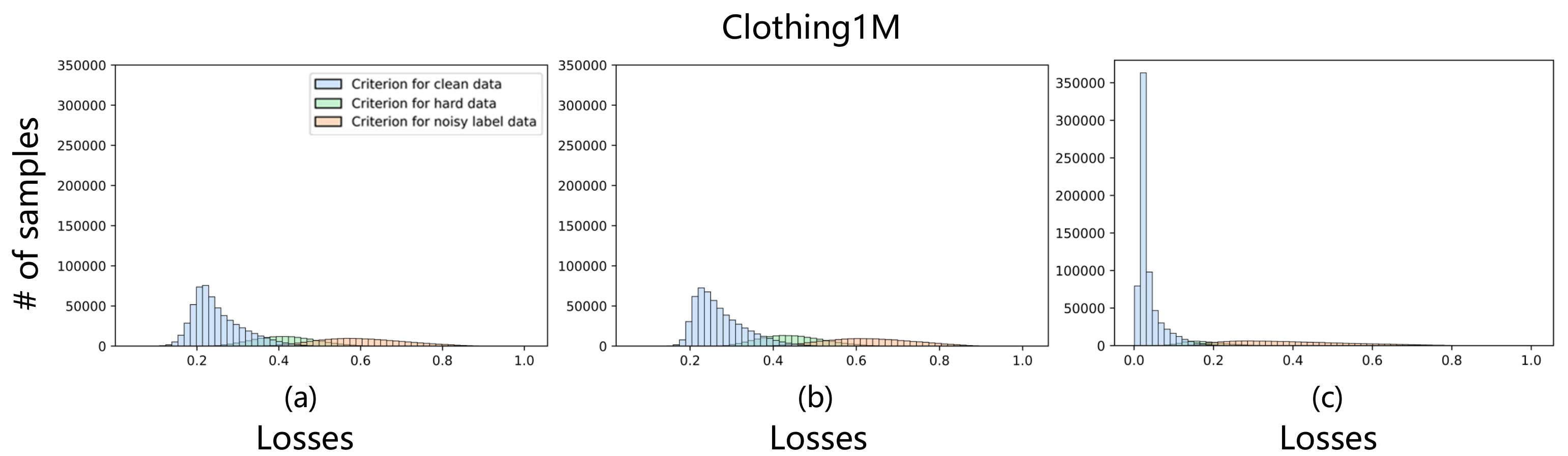}
	\end{center}
	\caption{The distribution of normalized losses of all samples from Clothing1M partitioned by our Tripartition criterion. The model is pre-trained on ImageNet. (a) The distribution at the end of the second epoch. (b) The distribution at the end of the fifth epoch. (c) The distribution of epochs at the optimal accuracy.}
	\label{fig:loss_clothing}
	\end{figure*}

	\cref{fig:loss1} shows that the population of hard subset is becoming smaller with the progressing of training. The test is carried out 3 times on CIFAR-100 with symmetric and realistic noise types and 50\% noise ratio.
	
	\begin{table*}[th]
		\centering
		\vspace{-0.06 in}
		\small
		\resizebox{145mm}{!}{
			\begin{tabular}{llc||llc||llc}
				\hline
				\multicolumn{2}{c|}{Similar classes}	&Similarity &\multicolumn{2}{c|}{Similar classes}	 & Similarity&\multicolumn{2}{c|}{Similar classes}	& Similarity\\ \hline	
				maple\_tree & oak\_tree     & 0.337 & poppy      & tulip     & 0.249 & beaver        & otter        & 0.222 \\
				dolphin     & whale         & 0.312 & bed        & couch     & 0.246 & lion          & tiger        & 0.220  \\
				plain       & sea           & 0.306 & poppy      & rose      & 0.245 & clock         & plate        & 0.220  \\
				bus         & pickup\_truck & 0.301 & snake      & worm      & 0.245 & pickup\_truck & tank         & 0.219 \\
				girl        & woman         & 0.298 & bottle     & can       & 0.242 & flatfish      & ray          & 0.217 \\
				bus         & streetcar     & 0.296 & apple      & pear      & 0.242 & plain         & road         & 0.216 \\
				bicycle     & motorcycle    & 0.275 & dolphin    & shark     & 0.237 & chimpanzee    & woman        & 0.216 \\
				baby        & girl          & 0.273 & baby       & boy       & 0.232 & shark         & trout        & 0.215 \\
				maple\_tree & willow\_tree  & 0.270  & poppy      & sunflower & 0.232 & lawn\_mower   & tractor      & 0.214 \\
				castle      & house         & 0.260  & bowl       & plate     & 0.231 & house         & streetcar    & 0.211 \\
				oak\_tree   & pine\_tree    & 0.258 & cloud      & plain     & 0.231 & leopard       & lion         & 0.210  \\
				cloud       & sea           & 0.257 & hamster    & mouse     & 0.230  & otter         & seal         & 0.210  \\
				oak\_tree   & willow\_tree  & 0.255 & chair      & couch     & 0.230  & beetle        & spider       & 0.209 \\
				orchid      & tulip         & 0.255 & bowl       & cup       & 0.229 & bed           & chair        & 0.209 \\
				beetle      & cockroach     & 0.254 & rose       & tulip     & 0.229 & orange        & pear         & 0.209 \\
				streetcar   & train         & 0.253 & apple      & orange    & 0.229 & maple\_tree   & rose         & 0.207 \\
				leopard     & tiger         & 0.253 & television & wardrobe  & 0.227 & pickup\_truck & tractor      & 0.205 \\
				man         & woman         & 0.251 & keyboard   & telephone & 0.227 & forest        & willow\_tree & 0.205 \\
				orange      & sweet\_pepper & 0.250  & boy        & girl      & 0.224 & crab          & lobster      & 0.205 \\
				apple       & sweet\_pepper & 0.250  & mouse      & shrew     & 0.223 & mountain      & sea          & 0.204  \\\hline
		\end{tabular}}
		\vspace{- 0.1 in}
		\caption{The top 60 similar class-pairs in descending order from CIFAR-100. The number is the cosine similarity.}
		\label{tab:class}
		\vspace{- 0.15 in}
	\end{table*}

	%
	%

	
	\section{The effectiveness of Tripartition criterion.}
	To demonstrate the effectiveness of our proposed Tripartition criterion, we conduct a control experiment for Small-loss based and GMM based methods. Specifically, we replace the training data by the one partitioned by our Tripartition criterion for each method, and apply their own learning strategies to verify if their performances would be improved. We choose three versions of a network (PreActResNet-18) which was trained in the early, middle, and late training stages on CIFAR-100 with 50\% realistic noise. These three versions achieve accuracies of 31.38\%, 41.93\%, and 63.50\%, respectively.
	Taking the version trained in the middle stage (41.93\%) as an example, the training data are divided by this version using three partition criteria, i.e. Small-loss (the partition criterion used by Co-teaching), GMM (the partition criterion used by DivideMix), Tripartition, respectively. We then have three different partitioned training datasets for the control experiment. For Co-teaching, it is further trained by the dataset divided by the Small-loss criterion. During the training, the dataset is not divided anymore. The control group is trained by the dataset divided by our Tripartition criterion. Two networks will converge after about 200 epochs. Their classification accuracies are listed in \cref{tab:effect}. One can see the network trained by dataset divided by our Tripartition criterion has 1.7\% performance gain. For DivideMix, we follow the same protocol. The network is trained by the datasets divided by the GMM criterion and Tripartition criterion, respectively. The results show that the network trained by dataset divided by our Tripartition criterion achieves 1.04\% performance gain.
	
	\begin{table}[h]
		\centering
		\small
		\resizebox{\linewidth}{!}{\begin{tabular}{c|c|cc}
				\hline
				\multicolumn{1}{c|}{Base\_Model (Acc.\%)} & \multicolumn{1}{c|}{Method}    & Partition criterion & Acc.(\%) \\
				\hline
				\multirow{4}{*}{Early(31.38)}       & \multirow{2}{*}{Co-teaching} & Small-loss         & 26.52        \\
				&                               & Tripartition        & \textbf{27.20}        \\\cline{2-4}
				& \multirow{2}{*}{DivideMix}    & GMM                 & 33.85        \\
				&                               & Tripartition        & \textbf{34.08}        \\
				\hline
				\multirow{4}{*}{Middle(41.93)}        & \multirow{2}{*}{Co-teaching} & Small-loss         & 32.70    \\
				&                               & Tripartition        & \textbf{34.40}    \\\cline{2-4}
				& \multirow{2}{*}{DivideMix}    & GMM                 & 41.26    \\
				&                               & Tripartition        & \textbf{42.30}    \\
				\hline
				\multirow{4}{*}{Late(63.50)}      & \multirow{2}{*}{Co-teaching} & Small-loss         & 37.92    \\
				
				&                               & Tripartition        & \textbf{38.54}    \\\cline{2-4}
				& \multirow{2}{*}{DivideMix}    & GMM                 & 49.45    \\
				&                               & Tripartition        & \textbf{50.15}   \\
				\hline
		\end{tabular}}
		\vspace{-0.1 in}
		\caption{The effectiveness of Tripartition criterion on other models when they are in different training stages.}
		\label{tab:effect}
	\end{table}
	
	\begin{table*}[!ht]
		\centering
		\fontsize{6}{7}\selectfont
		\resizebox{\linewidth}{!}{
			\begin{tabular}{c|c|cllll||cllll}
				\hline
				\multicolumn{1}{l|}{Hyperparameter} & \multicolumn{1}{l|}{Dataset} & \multicolumn{5}{c||}{CIFAR-10}                                      & \multicolumn{5}{c}{CIFAR-100}                                     \\\hline
				\multirow{4}{*}{$\lambda_n$}                 & \multirow{2}{*}{Realistic}  & \multicolumn{1}{l}{10\%} & 20\% & 30\%        & 40\%       & 50\% & \multicolumn{1}{l}{10\%} & 20\% & 30\%        & 40\%       & 50\% \\\cline{3-12}
				&                             & \multicolumn{1}{l}{1}    & 1    & 1           & 1          & 1    & \multicolumn{1}{l}{10}   & 30   & 30          & 60         & 60   \\\cline{2-12}
				& \multirow{2}{*}{Symmetric}  & \multicolumn{2}{c}{20\%}        & \multicolumn{2}{c}{50\%} & 80\% & \multicolumn{2}{c}{20\%}        & \multicolumn{2}{c}{50\%} & 80\% \\\cline{3-12}
				&                             & \multicolumn{2}{c}{1}           & \multicolumn{2}{c}{1}    & 1    & \multicolumn{2}{c}{10}          & \multicolumn{2}{c}{60}   & 120 \\
				\hline
		\end{tabular}}
		\caption{The selection of $\lambda _n$ in varied noise setting.}
		\label{tab:lam}
	\end{table*}
	We can observe similar results of the version trained in the early stage (31.38\%) and the late stage (63.50\%). Our Tripartition criterion improves Co-teaching and DivideMix about 0.72\% and 0.23\%, 0.6\% and 0.7\%, respectively. Above experiments demonstrate our Tripartition does reach a better training data partition and help a model to achieve a better performance.

	\section{The selection of $\lambda_n$ for noisy subset}
\begin{table}[h]
		\vspace{-3.5mm}
			\fontsize{8}{9}\selectfont
		\centering
		\begin{tabular}{l|c|c||c|c}
			\hline
			\multirow{2}{*}{\diagbox [width=8em,trim=r] {Strategy}{Noise type}} & \multicolumn{2}{c||}{Symmetric}   & \multicolumn{2}{c}{Realistic}   \\\cline{2-5}
			& 20\%           & 50\%           & 20\%           & 50\%           \\\hline
			$\lambda_n=1$   & 78.21          & 71.95         & 78.11          & 68.08          \\
			$\lambda_n=10$  & \textbf{78.65} & 72.95          & 78.21          & 68.30          \\
			$\lambda_n=40$  & 78.45          & 74.01          & \textbf{78.74} & 68.45          \\
			$\lambda_n=60$  & 78.31           & \textbf{74.73} & 78.24          & \textbf{68.62} \\
			$\lambda_n=100$ & 78.23          & 72.79          & 77.42          & 67.80     \\
			\hline
		\end{tabular}
		\caption{Evaluation of $\lambda_n$ for training strategy of noisy subset in CIFAR-100.}
		\label{tab:ablation1}
	\end{table}
	The $\lambda _n$ is a parameter to control the contribution of noisy subset. \cref{tab:lam} lists the selected values of $\lambda _n$ in varied noise settings on CIFAR datasets. The dataset with higher noise ratio and more complex data requires a larger $\lambda _n$ to learn sufficient information.  \cref{tab:ablation1} shows more detailed results with varied $\lambda _n$. One can see that Tripartite is not sensitive to $\lambda _n$, particularly on low noise ratio and realistic noise. On symmetric noise with a ratio of 50\%, the difference between the best and worst performances is less than 3\%.

	\section{Additional experiment results}
	\subsection{Results on CIFAR-10/CIFAR-100 with asymmetric noise}
	
	\cref{tab:asym} shows the test accuracy on CIFAR-10 and CIFAR-100 with 40\% asymmetric noise. (The noise ratio over 50\% is not used because some classes become theoretically indistinguishable in such setting.)
	
	\begin{table}[h!]
		\centering
		\small
		\begin{tabular}{l|c|c}
			\hline
			\multirow{2}{*}{\diagbox [width=14em,trim=l] {Method(Backbone)}{Dataset}}     & \multirow{2}{*}{\textbf{CIFAR-10}} & \multirow{2}{*}{\textbf{CIFAR-100}} \\
			& & \\
			\hline
			Cross entropy(ResNet34)     & 80.11             & 42.74                                  \\
			Bootstrap(ResNet34)         & 81.21             & 45.12                                  \\
			Forward(ResNet34)           & 83.55             & 34.44                                  \\
			GSE(ResNet34)               & 76.74             & 47.22                                  \\
			SL(ResNet34)                & 82.51             & 69.32                                  \\
			DivideMix(PreActResNet-18)  & \underline{93.40}             & 72.10                                   \\
			ELR+(PreActResNet-18)       & 92.70              & \underline{76.50}                                   \\
			Tripartite(PreActResNet-18) & \textbf{94.01}             & \textbf{76.89}              \\
			\hline
		\end{tabular}
		\caption{Accuracy comparisons on CIFAR-10 and CIFAR-100 with 40\% asymmetric noise.}
		\label{tab:asym}
	\end{table}
	\subsection{Additional results of Tripartition criterion}
	\cref{tab:sym} and \cref{tab:real} are the numerical versions of Fig. 6 in the paper.
	
	\section{Dataset details}
	The details of datasets used in the paper are shown in \cref{tab:detail}.
	
	\begin{table}[h]
		\centering
		\resizebox{\linewidth}{!}{
			\begin{tabular}{lcccc}
				\hline
				Datasets     & \# of training & \# of test & \# of class & size     \\
				\hline
				CIFAR-10     & 50,000         & 10,000     & 10          & 32x32    \\
				CIFAR-100    & 50,000         & 10,000     & 100         & 32x32    \\
				Food-101N     & 310,000        & 5,000      & 101         & 64x64    \\
				Clothing1M   & 1,000,000      & 10,526     & 14          & * \\
				WebVision1.0 & 2,400,000      & 50,000     & 1000        & *\\
				\hline
		\end{tabular}}
		\vspace{-0.04 in}
		\caption{The detailed configurations of five benchmark datasets in the paper.}
		\label{tab:detail}
	\end{table}
	
	\begin{table*}[p]
		\centering	
		\small
		\begin{tabular}{ccccccc}
			\hline
			\multicolumn{1}{c}{\multirow{2}{*}{Epoch}} & \multicolumn{1}{c}{\multirow{2}{*}{Acc.(\%)}} & \multicolumn{2}{c}{Clean\_subset} & \multicolumn{2}{c}{Noisy\_subset} & Hard\_subset \\
			\cline{3-7}
			\multicolumn{1}{c}{}                       & \multicolumn{1}{c}{}                      & Partition acc.     & Total nums    & Partition acc.     & Total nums    & Total nums   \\
			\hline
			30                                         & 44.06                                    & 90.19             & 14873         & 83.65             & 23413         & 11714        \\
			50                                         & 46.49                                    & 97.79             & 17421         & 82.32             & 29543         & 3036         \\
			100                                        & 51.67                                    & 97.96             & 18901         & 85.32             & 28501         & 2598         \\
			150                                        & 53.71                                    & 97.75             & 19833         & 87.15             & 27765         & 2402         \\
			200                                        & 68.57                                    & 97.71             & 22509         & 90.83             & 26212         & 1279         \\
			250                                        & 73.23                                    & 97.84             & 22663         & 91.26             & 26030         & 1307         \\
			300                                        & 73.68                                    & 97.97             & 23132         & 92.87             & 25329         & 1539  \\
			\hline
		\end{tabular}
		\caption{The results of Tripartition on CIFAR-100 with symmetric noise type and 50\% noise ratio.}
		\label{tab:sym}
	\end{table*}

	\begin{table*}[p]
		\centering
		\small
		\begin{tabular}{cccccccc}
			\hline
			\multicolumn{1}{c}{\multirow{2}{*}{Epoch}} & \multicolumn{1}{c}{\multirow{2}{*}{Acc.(\%)}} & \multicolumn{2}{c}{Clean subset} & \multicolumn{2}{c}{Noisy subset} & Hard subset \\
			\cline{3-7}
			\multicolumn{1}{c}{}                       & \multicolumn{1}{c}{}                      & Partition acc.     & Total nums    & Partition acc.     & Total nums    & Total nums   \\
			\hline
			30    & 39.20  & 68.48 & 17456 & 72.69 & 15485 & 17059 \\
			50    & 45.02 & 83.14 & 18801 & 76.11 & 25730 & 5469  \\
			100   & 46.89 & 83.78 & 20026 & 78.17 & 25390 & 4584  \\
			150   & 48.17 & 85.91 & 20361 & 80.20  & 25358 & 4281  \\
			200   & 67.10  & 81.76 & 24738 & 83.53 & 22896 & 2366  \\
			250   & 68.01 & 80.71 & 25646 & 84.07 & 22416 & 1938  \\
			300   & 67.76 & 80.21 & 25952 & 84.16 & 22288 & 1760 \\
			\hline
		\end{tabular}
		\caption{The results of Tripartition on CIFAR-100 with realistic noise type and 50\% noise ratio.}
		\label{tab:real}
	\end{table*}
	{\small
		\bibliographystyle{ieee_fullname}
		\bibliography{reference}
	}